\documentclass{article}

\usepackage[final]{neurips_2025}

\usepackage[utf8]{inputenc} % allow utf-8 input
\usepackage[T1]{fontenc}    % use 8-bit T1 fonts
\usepackage{hyperref}       % hyperlinks
\usepackage{url}            % simple URL typesetting
\usepackage{booktabs}       % professional-quality tables
\usepackage{amsfonts}       % blackboard math symbols
\usepackage{nicefrac}       % compact symbols for 1/2, etc.
\usepackage{microtype}      % microtypography
\usepackage{xcolor}         % colors
\usepackage{xspace}
\usepackage{graphicx}
\usepackage{wrapfig}
\usepackage{subfigure}
\usepackage{amsmath}
\usepackage{multirow}
\usepackage{listings}
\usepackage{pifont}
\usepackage{tabularx}
\usepackage{makecell}
\usepackage[table]{xcolor}
\usepackage{wrapfig}
\usepackage{subcaption}
\usepackage{float}

\newcommand{\ie}{\emph{i.e.,}\xspace}

\newcommand{\eg}{\emph{e.g.,}\xspace}

\newcommand{\ignore}[1]{}

\title{Towards General Continuous Memory for Vision-Language Models}
\author{%
  Wenyi Wu\thanks{Equal Contribution} ,~
  Zixuan Song$^*$,~
  Kun Zhou\thanks{Corresponding Author} ,
  Yifei Shao,
  Zhiting Hu,
  Biwei Huang
  \\
  University of California, San Diego.\\  % \\
  ~\texttt{kuzhou@ucsd.edu}\\
}

\begin{document}

\maketitle

%\vspace{-8mm}
%\begin{center}
%\footnotesize{$^{*}$Equal contribution}
%\end{center}

\maketitle

\begin{abstract}
Language models~(LMs) and their extension, vision-language models~(VLMs), have achieved remarkable performance across various tasks. However, they still struggle with complex reasoning tasks that require multimodal or multilingual real-world knowledge. To support such capabilities, an external memory system that can efficiently provide relevant multimodal information is essential. Existing approaches generally concatenate image and text tokens into a long sequence as memory, which, however, may drastically increase context length and even degrade performance. In contrast, we propose using continuous memory-a compact set of dense embeddings-to more effectively and efficiently represent multimodal and multilingual knowledge. Our key insight is that a VLM can serve as its own continuous memory encoder. We empirically show that this design improves performance on complex multimodal reasoning tasks. Building on this, we introduce a data-efficient and parameter-efficient method to fine-tune the VLM into a memory encoder, requiring only 1.2\% of the model’s parameters and a small corpus of 15.6K self-synthesized samples. Our approach CoMEM utilizes VLM's original capabilities to encode arbitrary multimodal and multilingual knowledge into just 8 continuous embeddings. Since the inference-time VLM remains frozen, our memory module is plug-and-play and can be flexibly integrated as needed. Extensive experiments across eight multimodal reasoning benchmarks demonstrate the effectiveness of our approach. Code and data is publicly released here \url{https://github.com/WenyiWU0111/CoMEM}.
\end{abstract}

% \begin{figure}[h]
%     \centering
%     \includegraphics[width=0.6\linewidth]{plots/Introduction.drawio.png}
%     \caption{An overview of the CoMEM architecture in comparison to the traditional RAG method.}
%     \label{fig:enter-label}
% \end{figure}

\begin{wrapfigure}[9]{r}{0.45\textwidth}
\vspace{-18pt}
    \centering
\includegraphics[width=0.45\textwidth]{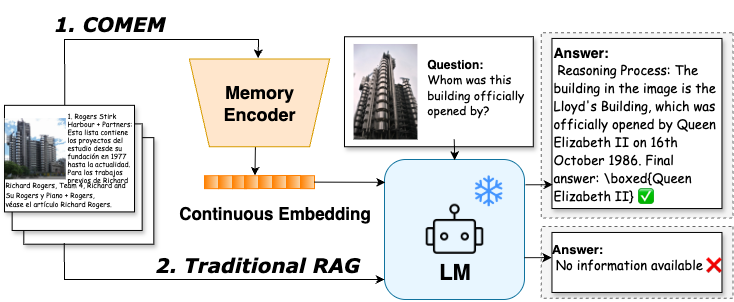}
    \caption{CoMEM architecture in comparison to the traditional RAG method.}
    \label{fig:enter-label}
\end{wrapfigure}

\section{Introduction}
\label{sec:intro}
Through large-scale training, language models~(LMs)~\citep{zhao2023survey,brown2020language} have demonstrated remarkable performance across diverse real-world tasks. LMs even surpass human capabilities in language reasoning tasks~\citep{dasgupta2022language} such as mathematical problem solving~\citep{ahn2024large}, commonsense reasoning\citep{wei2022chain}, and code synthesis~\citep{jiang2024survey}.
However, when confronted with complex reasoning tasks that demand multimodal or multilingual world knowledge, both LMs and their vision-language model (VLM) extensions continue to face significant challenges~\citep{li2023evaluating,wen2025token}, primarily due to insufficient world knowledge representation.

Inspired by how humans offload facts, plans, and ideas to external repositories like notebooks or databases for on-demand access, it is promising to develop a general external memory\footnote{In this paper, external memory denotes any detachable module or function that supplies the knowledge without changing LM parameters, in contrast to internal memory that embeds knowledge by modifying parameters.} that contains useful world knowledge for augmenting VLMs~\citep{lewis2020rag, wang2023longmem}.
Early approaches directly concatenate the collected useful information into a long token sequence, and feeds it into VLMs~\citep{lewis2020rag,guu2020realm} \eg retrieval-augmented generation~(RAG) methods.
However, multimodal representations demand significantly more input tokens (\eg 8 to 11427 tokens per image in Qwen2.5-VL~\citep{bai2025qwen2.5-vl}).
Thus, simple concatenation would greatly increase the input length, making it difficult for the memory content to be used~\citep{yu2024visrag} (see the degradation in performance shown in Table~\ref{tab:empirical} after using RAG).
To solve the token overload issue, token pruning methods have been proposed to remove unimportant in-context tokens~\citep{chen2024image, zhang2024sparsevlm}.
However, token pruning generally leads to incomplete contextual contents, which impedes the VLM's ability to accurately understand and utilize the compressed information~\citep{wen2025token}.

Compared to discrete tokens, continuous embeddings naturally have stronger representation capability for complex data~\citep{elhage2021mathematical,olsson2022incontext, hao2024training}.
This advantage makes them particularly promising for memory encoder architectures designed to condense multimodal information into continuous representations.
However, training such encoders faces two key challenges: (1) achieving generalizable compression ability across diverse multimodal inputs, and (2) maintaining semantic alignment with the VLM~\citep{jaegle2021perceiver}.
While large-scale training can improve performance, it greatly increases the training cost and becomes heavily sensitive to the training data distribution. 
For example, when dominated by simple cases or a single domain, the encoder tends to overfit and degrade generalization performance~\citep{arpit2017closer}~\citep{grangier-iter-2022-trade}.

%the effective compression of multimodal information is the key for the external memory of VLMs.
%learns a memory encoder to condense the information into continuous embeddings~\citep{shi2024compressing}.
%Although the continuous embeddings based memory is promising to solve this issue, 

%Inspired by recent findings on LLM representation transferability, we investigate whether a pretrained Vision-Language Model (VLM) can serve as a memory encoder for itself without any additional training. Our empirical study confirms this hypothesis: a small number of key token representations—extracted from intermediate layers of the VLM and selected by attention scores—are sufficient to preserve the essential semantic information of each instance. These continuous and compact embeddings can be directly injected as memory vectors into the inference-time model as memory, improving VLM's performance in knowledge-intensive multimodal tasks. 

%Therefore, it is still a critical research challenge to 

In this paper, we focus on efficiently training a general continuous memory encoder to effectively supply multimodal knowledge for VLMs.
To avoid costly training for semantic alignment, it is essential to minimize the representation gap between the memory encoder and downstream VLMs before training.
Therefore, a natural way is to use the VLM itself as the memory encoder.
%initialize an encoder that with the downstream LLM during inference. 
%For efficient memory encoder training, 
Our empirical study confirms that the VLM can serve as a memory encoder for itself without any additional training. 
Benefiting from the stacked self-attention mechanism, its generated continuous embeddings in each layer have already aggregated rich semantic information~\citep{vaswani2017attention}~\citep{clark2019bert}.
As shown in Fig.~\ref{fig:attn_token_selection}, even a simple rule-based embedding selection strategy for constructing the memory can greatly boost the performance of VLMs in complex multimodal multilingual reasoning tasks, compared to RAGs.

%The continuous, compact embeddings extracted from its intermediate layers contain rich semantic information and can be seamlessly utilized by the inference-time model as memory.

% For Efficient Memory Encoder Training, we aim to find a well-initialized encoder with smallest gap with the LLM for inference. Inspired by existing findings about LLM representation transferability, we study if the VLM can be the memory encoder for itself, without any training. Our experimental results find that few key token representations in the VLM are able to keep the key information of the instance, which can serve as the memory for improving downstream tasks.

Based on our empirical findings, we propose a data-efficient and parameter-efficient training recipe to further improve the compressor rate and adaptation performance of the VLM-based general continuous memory encoder.
Concretely, we only need to fine-tune the low-rank adaptation matrices~(LoRA)~\citep{hu2021lora} in the VLM-based memory encoder, and a lightweight Q-Former~\citep{li2023blip2} for further compressing the VLM representations into only eight embeddings, 1.2\% parameters in total.
In terms of data, we only need the VLM itself to synthesize 15.6k samples for training.
This efficient training strategy enables our continuous memory to reuse the original ability of the VLM, to effectively encode multimodal and multilingual knowledge.
Since we do not need to train the inference VLM, our memory is plug-and-play and can be flexibly integrated with the VLM when necessary.

To demonstrate the effectiveness of our approach, we apply our method to state-of-the-art VLMs, and evaluate the performance across eight visual reasoning benchmarks. For six English visual reasoning benchmarks, our method achieves an average improvement of +8.0\% on Qwen2-Instruct-VL and +7.7\% on Qwen2.5-Instruct-VL. 
On two multilingual multimodal benchmarks, our approach further improves performance by +5.1\% and +4.3\% on Qwen2-Instruct-VL and Qwen2.5-Instruct-VL, respectively.
Furthermore, our adaptation study results also indicate the transferability of our VLM-based memory encoder to improve LMs in visual reasoning tasks.
The long context understanding study also exhibits the stable and superior performance of our method.
%These extensive experimental results demonstrate our method's generalizability, robustness, and significant improvement on SOTA VLMs and LLMs.

% \Biwei{write a paragraph for contributions}

\newcommand{\cmark}{\ding{51}}  % ✓
\newcommand{\xmark}{\ding{55}}  % ✗

\begin{table}[h]
\caption{Comparison between our approach and other representative line of work.}
\centering
\small
\setlength{\tabcolsep}{0.7pt}
\resizebox{\textwidth}{!}{%
\begin{tabular}{llcccccc}
\toprule
 & & \multicolumn{2}{c}{\textbf{Properties}} & \multicolumn{2}{c}{\textbf{Scenarios}} & \multicolumn{2}{c}{\textbf{Training Cost}} \\
 \cmidrule(lr){3-4} \cmidrule(lr){5-6} \cmidrule(lr){7-8}
\textbf{Category~~} & \textbf{Method} & Continuous~ & ~Pulg-and-Play~ & Multimodal & Multilingual  & Data & Parameters \\
\midrule
\multirow{3}{*}{Multimodal-RAG}& EchoSight~\cite{echosight} &
\xmark &
\cmark &
\xmark Text &
\xmark &
900K &
300M
\\
& ReflectiVA~\cite{reflectiva} &
\xmark &
\xmark &
\cmark Image+Text &
\xmark &
6.82M &
8B
 \\
& RoRA-VLM~\cite{roravlm} &
\xmark &
\xmark &
\cmark Image+Text &
\xmark &
1M &
7B
 \\
\midrule
\multirow{3}{*}{Context-Compression~~}& xRAG~\cite{xrag} &
\cmark &
\cmark &
\xmark Text &
\xmark &
3M &
40M
\\
& KV-Distill~\cite{Chari2025KVDistillNL} &
\xmark &
\cmark &
\xmark Text &
\xmark &
500K+
 &150M

 \\
& VoCo-LLaMA~\cite{vocolama} &
\cmark &
\xmark &
\cmark Image/Video &
\xmark &
665K &
7B
 \\
\midrule
\multirow{4}{*}{LM Memory} & LONGMEM~\cite{wang2023longmem} &
\cmark &
\cmark &
\xmark Text &
\xmark &
114M&
558M
\\

& MA-LMM~\cite{he2024malmm} &
\cmark &
\cmark &
\cmark Video+Text &
\xmark &
NA&
200M
 \\
& M+ ~\cite{wang2025mplus}&
\cmark &
\xmark &
\xmark Text &
\xmark &
5M&
NA\\
& MemGPT ~\cite{memgpt2024packer}&
\xmark &
\cmark &
\xmark Text &
\xmark &
NA
&NA
 \\
 \hline
  \hline
& CoMEM &
\cmark &
\cmark &
\cmark Image+Text &
\cmark &
15K &
200M
 \\
\bottomrule
\end{tabular}
}
\label{tab:related_work_comparison_intro}
\end{table}

\section{Empirical Analysis with VLM as Memory Encoder}

% 1. RAG has bad performance on long and complex context (->compression)
% 2. Naive token selection performance is not stable (->model adaptation)
In this section, we conduct an empirical study to examine 
(1) whether VLM can serve as a continuous memory encoder to compress multimodal information into compatible embeddings, and
(2) whether a few embeddings from the VLM can preserve key information to improve multimodal reasoning tasks.
%a small number of key tokens can preserve sufficient information for downstream task, especially in knowledge intensive cases.

\subsection{Analysis Setup}
For the empirical study, we conduct experiments on two state-of-the-art VLMs, \ie Qwen2-VL-7B and Qwen2.5-VL-7B, and test the performance on three multimodal reasoning benchmarks.

\paragraph{Evaluation Settings.}
To compare the effectiveness of different memory and context compression methods, we select three benchmarks: InfoSeek~\citep{chen2023can}, OK-VQA~\citep{marino2019ok}, and A-OKVQA~\citep{schwenk2022okvqa}.
These benchmarks contain complex visual questions that require both accurate visual entity identification and multi-step reasoning to derive the correct answer.
Following existing work~\citep{caffagni2024wikillava}~\citep{roravlm}, for each question, we utilize CLIP-based retriever~\citep{radford2021learning} to collect relevant top-10 multimodal knowledge items from a Wikipedia-based source dataset WiT~\citep{srinivasan2021wit} to construct the input data for the memory.

% \begin{center}
\begin{table}[h]
\caption{Comparison of training-free memory methods. Bold indicates the best performance. For VLM-as-Memory here, we use the cache KV from a VLM without fine-tuning, which differs from the main method described in Section \ref{sec:approach} and is intended for preliminary exploration.}
% and underlined values denote the second-best.}
\centering
\small
% \resizebox{\textwidth}{!}{%
\begin{tabular}{llcccccc}
\toprule
& & \multicolumn{3}{c}{\textbf{InfoSeek}} & \multirow{2}{*}{\textbf{OKVQA}} & \multirow{2}{*}{\textbf{AOKVQA}} \\
\cmidrule(lr){3-5}
 \textbf{Backbone Model} &\textbf{Method}& Query & Entity & All & & \\
\midrule
% \rowcolor{gray!20}
\multirow{5}{*}{\textbf{Qwen2.5-VL-Instruct}} 
&- & 22.5 & 22.4 & 22.5 & 35.0 & 39.8 \\
&+RAG & 17.7 & 18.8 & 18.2 & 31.3 & 34.9 \\
&+FastV & 26.2 & 22.6 & 24.2 & 31.5 & 34.9 \\
&+VLM-as-Memory & 29.3 & \textbf{28.0} &\textbf{28.6} & 37.3 & \textbf{44.4} \\
%&+VLM-as-Memory+Random &28.7&24.9&26.6&34.6&38.4\\
%&+VLM-as-Memory+Pooling &27.3&25.3&26.2&25.6&26.4\\
&+VLM-as-Memory+AS &\textbf{30.0}&25.3&27.5&\textbf{37.9}&41.8\\
% \midrule
% \rowcolor{gray!20}
\midrule
\midrule
\multirow{5}{*}{\textbf{Qwen2-VL-Instruct}}
&- & 17.9 & 17.8 & 17.9 & 36.3 & 41.8 \\
&+RAG & 22.7 & 19.0 & 20.5 & 41.9 & 45.3 \\
&+FastV & 23.6 & 23.8 & 23.7 & \textbf{42.0} & \textbf{45.4} \\
&+VLM-as-Memory & 28.8 & \textbf{29.7} & 29.3 & 37.7 & 38.9 \\
%&+VLM-as-Memory+Random &26.9&28.4&27.6&33.1&35.3\\
%&+VLM-as-Memory+Pooling &22.5&27.4&24.7&27.2&34.9\\
&+VLM-as-Memory+AS &\textbf{31.7}&28.8&\textbf{30.2}&34.3&36.4\\
\bottomrule
\end{tabular}%
% }
\label{tab:empirical}
\end{table}
% \end{center}

%retrieval-augmented generation task, where SOTA VLMs Qwen2 and Qwen2.5 are required to generate answers of knowledge intensive vision questions with retrieved knowledge.
%We assess the performance of our continuous memory using three Knowledge-Intensive Visual Question Answering (VQA) benchmarks, which evaluate models on information-seeking, retrieval-augmented reasoning, and multimodal entity understanding tasks. These include InfoSeek\citep{chen2023can}, OK-VQA\citep{marino2019ok}, and A-OKVQA\citep{schwenk2022okvqa}. Each benchmark is selected to test specific aspects of our model's capabilities. Detailed description of these benchmarks are provided in Appendix~\ref{appendix:benchmarks}.

%mainly study the performance of open-source LVLMs Qwen2 and Qwen2.5 with training free continuous memory on multiple  like . 

\paragraph{Memory Methods.}
We test the effectiveness of our VLM-as-memory method by comparing with RAG, token pruning, and our variations using different embedding selection strategies.

$\bullet$ \emph{Vanilla RAG}: it simply concatenates all the multimodal knowledge items into a long sequence, and then feeds it with the visual question as the input of VLM.

$\bullet$ \emph{FastV}~\citep{chen2024image}: it adopts a token pruning strategy that discard the image tokens with lower attention scores from the multimodal knowledge items. Then, the pruned token sequence is fed into the VLM.

$\bullet$ \emph{VLM-as-Memory}: we utilize the VLM itself to encode the knowledge items, and extract the hidden states in all layers as the memory. These are concatenated at corresponding layers of the VLM. For efficiency, we only add the memory in 17-19 layers.

$\bullet$ \emph{VLM-as-Memory+Attn}: we utilize the average attention scores across all layers within the VLM to select the top-25\% key continuous embeddings to compose the memory.

\subsection{Results and Findings}
In this part, we present the results and discuss the findings to analyze whether VLM can be a memory encoder, assessing both their compression efficiency and semantic alignment capabilities.

\begin{wrapfigure}[14]{r}{0.38\textwidth}
\vspace{-18pt}
    \centering
    \includegraphics[width=0.38\textwidth]{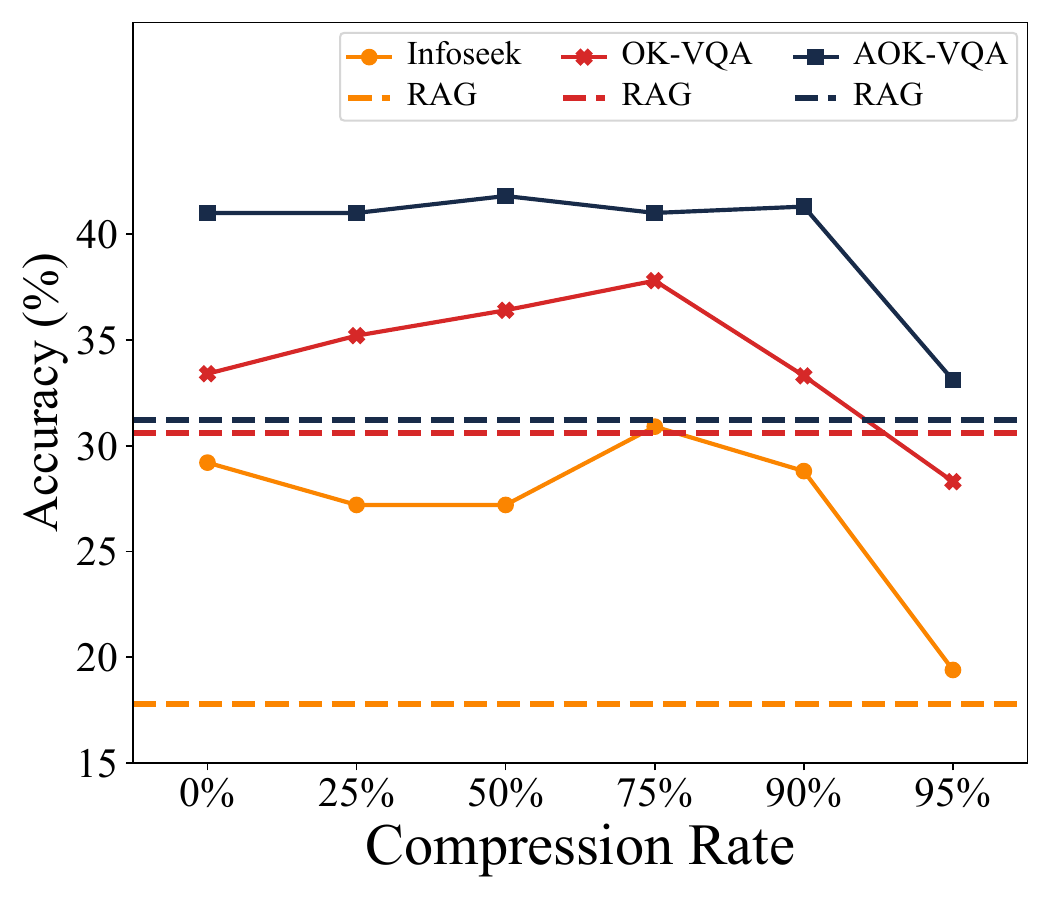}
    \caption{Qwen2.5 accuracy with varying attention-based compression rates.}
    \label{fig:attn_token_selection}
\end{wrapfigure}

\paragraph{Effectiveness Study of VLM-as-Memory Methods.}
As shown in Table~\ref{tab:empirical}, the vanilla RAG method causes performance degradation in several tasks, underscoring its limitations for memory integration.
% due to its produced rather long memory for the VLM.
%will lead to a very long input, which greatly increases the difficulty of using the memory.
FastV mitigates this issue by pruning redundant tokens in the memory, resulting in measurable improvements.
Notably, the VLM-as-Memory method outperforms both approaches across most tasks, suggesting that the VLM’s continuous embeddings are inherently more compatible with its own processing than token-based input.
Furthermore, with the addition of a simple attention-based compression mechanism, the VLM-as-Memory method achieves even greater performance gains.
Thus, we conclude that:

\emph{\textbf{(1) VLMs can effectively serve as their own memory encoders for external multimodal knowledge.}} The continuous embeddings they generate can be directly reused by the same model without requiring additional training.

\emph{\textbf{(2) The continuous embeddings produced by VLMs effectively preserve knowledge content.}} They remain robust under simple compression strategies and reliably enhance performance.

\paragraph{Compressibility Study of VLM-as-Memory Methods.}

To study the compressibility, another key feature of an effective memory mechanism, we investigate how performance varies under different compression rates using the VLM-as-Memory approach.
As shown in Fig.~\ref{fig:attn_token_selection}, despite employing a simple token selection strategy, our method outperforms the baseline even at a high compression rate of 5%.
This suggests that a small number of continuous embeddings already encapsulate most of the essential contextual information from the input.
Therefore, we can conclude that: 

\emph{\textbf{(3) The continuous embeddings produced by VLMs support high compression rates.}} This highlights the potential for achieving even greater compression through more advanced compression methods.

%investigate the effectiveness of continuous memory representations, and potential problems of token selection and pooling strategies, we evaluate how model performance varies with different percentages of top-ranked memory tokens selected based on attention scores, and different number of pooling tokens (Plots\ref{tab:performance_comparison_qwen2.5} and \ref{tab:performance_comparison_qwen2}). These line plots  reveal two findings: 

%$\bullet$ \textbf{Only a small subset of key memory representations are effective}. Retaining just the top 25\% of memory tokens—either by attention-based selection or by pooling—already yields significant improvements over both the base model and RAG, with attention-based selection outperforming pooling. This indicates the potential of distilling rich and complex semantic information into a compact set of representations.

%$\bullet$ \textbf{A learnable aggregation method is needed for better and more stable performance.} 
%Determining the optimal proportion of tokens to retain is a difficult hyperparameter choice, and performance can vary significantly across different benchmarks when using fixed, training-free strategies like attention-based selection or pooling. This sensitivity limits the robustness and generalizability of such methods. To overcome these challenges, it is essential to develop a learnable token aggregation mechanism that can adaptively compress and preserve rich semantic information.

\section{Approach}
\label{sec:approach}
\begin{figure}[h]
    \centering
    \includegraphics[width=1\linewidth]{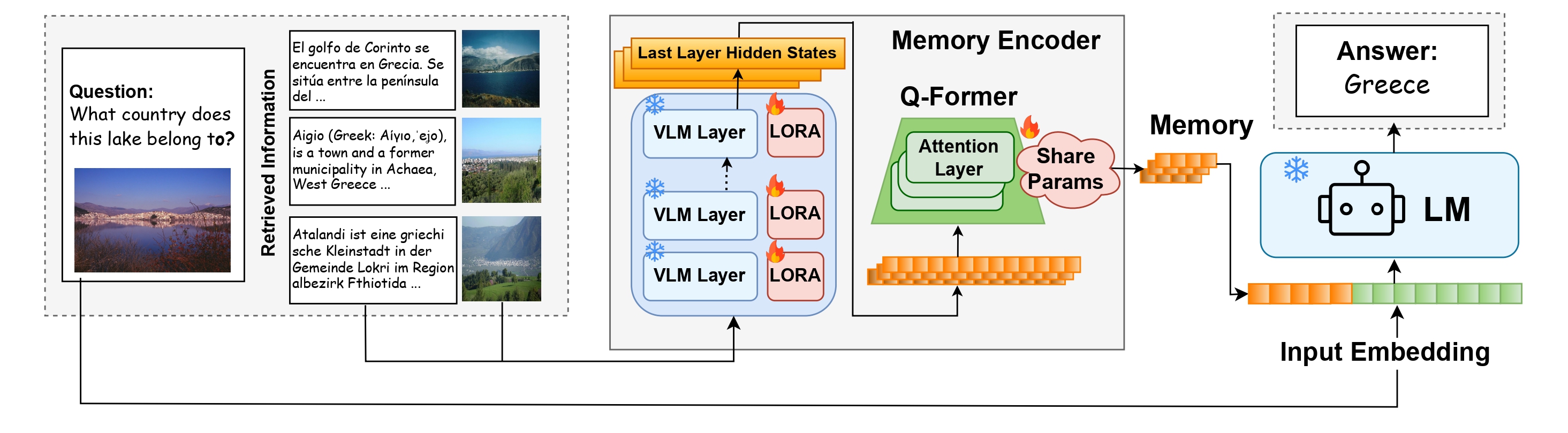}
    \caption{Overview of the CoMEM architecture. Given a vision-language query, the system retrieves relevant multimodal knowledge via visual features. Retrieved image-text pairs are processed by a Memory Encoder—which consists of a VLM and Q-Former—to generate a dense continuous memory. This memory and the original query are fed into a frozen LM to produce accurate, grounded answers.}
    \label{fig:enter-label}
\end{figure}
According to our empirical study, the VLM can be an effective memory encoder for itself, owing to the satisfactory semantic alignment and compressibility of its produced continuous embeddings.
Building on this insight, we aim to efficiently train the VLM into a continuous memory encoder, to supply supplementary multimodal knowledge during inference.
%train a VLM-based memory encoder, using few data and 
%are not fully optimized. To address this, we want to leverage the VLM itself, using only minimal data and parameters—while keeping the inference model frozen. This ensures an efficient, plug-and-play solution adaptable to various scenarios.
Concretely, we add a trainable lightweight Q-Former to control the compression rate, synthesize a small training dataset using the VLM itself, and perform data-efficient and parameter-efficient training.

%Given this we propose a method (*), which employs the VLM and a lightweight Q-Former as a knowledge encoder. This setup enables the distillation of complex, multimodal, and multilingual long-context inputs into a compact memory representation of just 8 tokens. These dense memory vectors are then injected into inference-time VLMs—or even standard LLMs—for effective knowledge transfer.

\subsection{Task Definition.}
% Edit: use less formula
We aim to train a general-purpose continuous memory encoder capable of mapping arbitrary multimodal and multilingual data into continuous embeddings that augment the knowledge of a VLM.
To ensure plug-and-play compatibility, we keep the VLM's parameters frozen during inference, while continuous memory embeddings are directly used for downstream tasks.
To achieve this, the memory encoder should (1) efficiently condense diverse multimodal and multilingual data and (2) produce embeddings that are both readable and functional for the VLM.

%To derive the correct answer, the model should fulfill two key requirements: (1) possess strong retrieval capabilities to extract the most relevant factual information; and (2) effectively fuse multimodal long-context knowledge as memory and exploit it for downstream reasoning tasks. In the following sections, we demonstrate that our model exhibits robust multimodal fusion ability and achieves strong performance even when relying on a simple CLIP image-to-image retrieval approach.

In this paper, we focus on using general continuous memory to enhance VLMs in complex multimodal reasoning tasks. 
Formally, given an instance comprising an image $i$ and a natural language question $q$, the task is to predict an accurate answer $a$.
Following prior work~\citep{roravlm}, we assume access to relevant multimodal knowledge items (from external knowledge source), each consisting of an image $\tilde{i}$ and a natural language description $\tilde{d}$.
Our memory encoder learns to transform each knowledge item into a continuous vector, formulated as $\textbf{V}_t=f(\tilde{i}_t, \tilde{d}_t)$.
These vectors are aggregated into a unified memory, which the VLM then utilizes for answer prediction: $p(a|i,q,\{\textbf{V}_t\}_{t=1}^{k})$.

\subsection{VLM-Based Continuous Memory}
For our continuous memory, the core idea is to leverage the VLM with a Q-Former as the encoder, and adopt a simple plug-and-play mechanism that enables the VLM to use the memory information.

%To effectively incorporate external multimodal and multilingual knowledge into a frozen language model, we propose a continuous memory framework which consists of two key components: a \textit{Content Compression Module}  and a \textit{Plug-and-Play Mechanism}. This design enables efficient knowledge injection without modifying or fine-tuning the backbone language model, while ensuring \textit{scalability, modularity, and cross-task reusability}.

\paragraph{Continuous Encoder.}
Given each multimodal knowledge item $\langle \tilde{i}_t, \tilde{d}_t \rangle$, we first use the VLM to encode it and collect the continuous representations $\textbf{E}_{t}$ in the last layers.
Then, we employ a query Transformer~(Q-Former) as the compressor to condense $\textbf{E}_{t}$ into $k$ continuous embeddings $\textbf{V}_t$.
The Q-Former consists of $k$ query embeddings $\mathbf{q}$ and $L$ Transformer layers.
In the first layer, the query embeddings attend to all the continuous representations from the VLM through the cross-attention mechanism. The output representations are then used as query embeddings for the next layer, and the final layer outputs serve as the memory vector $\textbf{V}_t$.
The whole process is formulated as:
\begin{equation}
\mathbf{H}^{(0)} = \mathbf{q}, \qquad
\mathbf{H}^{(\ell)}
   = \mathrm{TransformerLayer}^{(\ell)}
     \bigl(\mathbf{H}^{(\ell-1)},\,
           \mathbf{E}_t\bigr),
\quad \textbf{V}_t=\mathbf{H}^{(L)}
\end{equation}

To reduce the parameter scale of the Q-Former, we share parameters across all Transformer layers, and set $k=8$.
In this way, only a few parameters are added, and any multimodal knowledge item will be compressed into 8 continuous embeddings.
This design ensures lower training cost and a higher compression rate\footnote{Notably, the average token number of knowledge items in this work is \textbf{643.7}, and few extremely long ones contain more than 2000 tokens. By compressing into 8 tokens, we can achieve more than \textbf{80$\times$} compression rate.}, which is helpful to handle large-scale knowledge items and save the storage cost.

%long contexts from each image and text pairs into a small number of dense memory tokens. The working mechanism of the Q-Former knowledge compressor is as follows:
%To encode the retrieved multimodal and multilingual knowledge into a compact and efficient form, we employ a Q-Former-based compression module. The Q-Former acts as an information compressor that distills long contexts from each image and text pairs into a small number of dense memory tokens. The working mechanism of the Q-Former knowledge compressor is as follows:
%(1) Each retrieved knowledge image and text pair is first encoded using a knowledge encoder VLM. 
%(2) The last layer hidden states of these retrieved knowledge are passed to a Q-Former module with 8 learnable query embeddings. 
%(3) The Q-Former cross-attends to the hidden states and outputs 8 task-agnostic memory tokens per context item.
%This design transforms a variable-length, high-dimensional input into a fixed-size, compressed representation. The output memory tokens are concatenated across all retrieved items and stored as the continuous memory \( M \in \mathbb{R}^{n \times d} \), where \( n = 8k \) for \( k \) retrieved items.

\paragraph{Plug-and-Play Mechanism.}
After obtaining the continuous embedding set $\{\textbf{V}_t\}_{t=1}^{n}$ for all multimodal knowledge items, we adopt a simple plug-and-play mechanism to equip the VLM the memory.
Concretely, we simply concatenate the embeddings into a sequence of $8\times n$ continuous vectors as the memory, which is prepended to the input embedding $\textbf{E}_{I}$ of the VLM during the inference time, formulated as $[\textbf{V}_1; \cdots; \textbf{V}_n, \textbf{E}_{I}]$.
In this way, the VLM can naturally perform autoregressive generation to predict the answer, using its originally learned knowledge and capabilities.

%A core feature of our approach is its plug-and-play design, which enables us to \textit{seamlessly inject external knowledge into any pretrained language model without modifying its architecture or parameters}. The working steps of the plug-and-play mechanism is as follows:
%(1) During inference, the generative model \( \mathcal{G} \) is kept \textbf{frozen} and no fine-tuning or adaptation is required.
%(2) The retrieved knowledge is encoded and compressed by a slightly fine-tuned vision-language encoder and Q-Former module, producing a knowledge dense memory \( M \).
%(3) The memory tokens \( M \) are \textbf{prepended} to the embedding of input image and query, and then passed directly to the inference model \( \mathcal{G} \).

%This modular setup ensures that the memory encoder can be reused across different LLMs with just a light-weight finetuning to bridge the gap between memory and inference model, making the system highly flexible and scalable.

\subsection{Efficient Training Recipe}
Since we introduce the Q-Former, we need to train its parameters to achieve full alignment between the continuous memory and the VLM.
Thanks to our design that employs the VLM as the memory encoder, this alignment can be efficiently accomplished through parameter-efficient training using only a small amount of self-synthetic multimodal and multilingual data.

\paragraph{Training Data Self-synthesis.}
To ensure training efficiency, we construct our training dataset by synthesizing responses using the VLM itself, based on multilingual and multimodal questions from existing benchmarks.
Specifically, we begin by selecting questions from the training sets of InfoSeek~\citep{chen2023can}, Encyclopedic-VQA (EVQA)~\citep{mensink2023encyclopedic}, and OK-VQA~\citep{marino2019ok} to ensure coverage of diverse multimodal reasoning tasks.
For each question, we retrieve three relevant image-text pairs from the WIT~\citep{srinivasan2021wit} knowledge base using CLIP, following the retrieval setup in prior work~\citep{roravlm}. These pairs serve as supplementary multimodal knowledge items.
We concatenate the question with knowledge items and input the sequence into Qwen2.5-VL-Instruct to simulate a vanilla RAG setting. Only outputs yielding correct answers are retained, resulting in 13.8k high-quality training instances.
To extend our dataset beyond English, we randomly select 200 training samples and employ GPT-4o-mini to translate the text part into nine languages: Bulgarian, Chinese, Egyptian Arabic, Filipino, French, Japanese, Portuguese, Russian, and Spanish. 
This results in an additional 1.8k multimodal multilingual training samples, which aims at activating our model's cross-lingual capabilities. 
In total, our final fine-tuning corpus for continuous memory includes 15.6K curated samples, covering a variety of multimodal tasks and languages.

%We construct the training data that contains the questions and related multimodal knowledge items
%To fully activate the potential of continuous memory, we construct a compact yet reliable corpus for model fine-tuning. The dataset is selected from three knowledge-intensive english visual question answering (VQA) datasets: InfoSeek, Encyclopedic-VQA (EVQA) and OK-VQA. For each question-image pair in their training datasets, we retrieve three relevant image-text pairs from the WIT knowledge base using CLIP, and incoporate them as external knowledge. We then process the augmented input (question, image, and retrieved knowledge) using the Qwen2-5-VL model and retain only those instances where the prediction matches the reference answer. We apply this filtering process on a small amount of original training data, and collect 3K EVQA, 9K InfoSeek, and 1.8K OK-VQA samples.

%Additionally, to extend our dataset beyond English, we randomly select 200 InfoSeek samples and employ GPT-4o mini to translate their questions and answers into nine languages: Bulgarian, Chinese, Egyptian Arabic, Filipino, French, Japanese, Portuguese, Russian, and Spanish, while keeping the original images unchanged. This results in an additional 1,800 multilingual training samples, which aims at activating our model's cross-lingual capabilities. In total, the final fine-tuning corpus consists of 13.8K english samples and 1.8K multilingual samples.

\paragraph{Parameter-efficient Fine-tuning.}
\label{sec:param-setting}
Given the above training data, we perform parameter-efficient fine-tuning on the Q-Former and LoRA layers in the VLM encoder.
For efficiency, we apply LoRA with a rank of 16 and share parameters across all layers of the Q-Former.
Therefore, only 1.2\% of total parameters are trainable.
The above parameter and data efficient designs guarantee that our entire training process can be completed on a single NVIDIA H100 GPU in 20 hours.
We also empirically find the training converges fast, and a single epoch is sufficient to achieve strong performance.
%Firstly, to efficiently adapt the knowledge encoder, we finetune the knowledge encoder VLM with LoRA~\citep{hu2021lora} using a small amount of synthetic instruction data. We apply LoRA with only 16 rank, modifying only 0.27\% of total parameters. This approach ensures that the majority of the model remains untouched, so that the semantic space of the memory and the inference-time LM is aligned. In parallel, we fully fine-tune a light weight Q-Former module with only 8 layers and parameters in each layer are the same so that only 180M parameters in total, ensuring the training efficiency and low computational cost.
%With this efficient setup, our training involves only a few million parameters and uses just \textbf{15k synthetic samples}, and the entire training process can be completed on a single NVIDIA H100 GPU, yet still achieves strong performance across 8 benchmarks. 

% These training samples reflect the model's own data distribution and are conditioned on retrieved knowledge, making them ideal for our light weight supervised finetuning.

\subsection{Discussion}

\label{subsec:comparison}
% Our proposed method, CoMEM, introduces a general and efficient continuous memory mechanism that can be seamlessly integrated into VLMs to provide relevant multimodal and multilingual knowledge. By leveraging the VLM itself as the memory encoder, our approach significantly enhances semantic alignment between the memory and the model, enabling flexible plug-and-play integration across a wide range of downstream tasks.

In Table~\ref{tab:related_work_comparison_intro}, we compare our method CoMEM with ten closely related works: \ie multimodal RAG (EchoSight~\cite{echosight} ReflectiVA~\cite{reflectiva}, and RoRA-VLM~\cite{roravlm}), context compression (xRAG~\citep{xrag}, KV-Distill~\citep{Chari2025KVDistillNL} and VoCo-LLaMA~\cite{vocolama}), and LLM memory methods (LONGMEM~\cite{wang2023longmem}, MA-LMM~\citep{he2024malmm}, M+~\citep{wang2025mplus}, and MemGPT~\citep{memgpt2024packer}). 
The comparison spans three dimensions: Properties, where we examine whether the method is continuous and plug-and-play; Scenarios, evaluating support for multimodal and multilingual inputs; and Training Cost, which includes the amount of training data required and trainable parameters.

% \begin{wraptable}[14]{r}{0.48\textwidth}
% \vspace{-11pt}  % Reduce top space
% \caption{Comparison of representative methods with our approach. "I", "T", and "V" denote Image, Text, and Video modalities, respectively.}
% \centering
% \small
% % \renewcommand{\arraystretch}{1.1}
% \setlength{\tabcolsep}{0.7pt}
% \begin{tabular}{lcccccc}
% \toprule
%  & \multicolumn{2}{c}{\textbf{Properties}} & \multicolumn{2}{c}{\textbf{Scenarios}} & \multicolumn{2}{c}{\textbf{Training Cost}} \\
%  \cmidrule(lr){2-3} \cmidrule(lr){4-5} \cmidrule(lr){6-7}
% \textbf{Method} & Con. & PP & MM & ML  & Data & Params \\
% \midrule
% EchoSight &
% \xmark &
% \cmark &
% \xmark T &
% \xmark &
% 900K &
% 300M
% \\

% ReflectiVA &
% \xmark &
% \xmark &
% \cmark I+T &
% \xmark &
% 6.82M &
% 8B
%  \\
% xRAG &
% \cmark &
% \cmark &
% \xmark T &
% \xmark &
% 3M &
% 40M
% \\

% VoCo-LLaMA &
% \cmark &
% \xmark &
% \cmark I/V &
% \xmark &
% 665K &
% 7B
%  \\
% LONGMEM &
% \cmark &
% \cmark &
% \xmark T &
% \xmark &
% 114M&
% 558M
% \\

% MA-LMM &
% \cmark &
% \cmark &
% \cmark V+T &
% \xmark &
% NA&
% 200M
%  \\
%  \hline
%   \hline
%  CoMEM &
% \cmark &
% \cmark &
% \cmark I+T &
% \cmark &
% 15K &
% 200M
%  \\
% \bottomrule
% \end{tabular}
% \label{tab:related_work_comparison_intro}
% % \vspace{-20pt}  % Reduce space at the bottom
% \end{wraptable}

While some existing methods also adopt continuous embeddings and support plug-and-play usage, they often require substantial training resources—typically involving millions of training samples and extensive parameter updates. In contrast, our method achieves comparable functionality with significantly reduced cost: it utilizes only 15.6k self-synthesized training samples and fine-tunes just 200M parameters, amounting to only 1.2\% of the full model.
Moreover, a key advantage of our method is our method can handle both multimodal (text and image) and multilingual data, which is very helpful for potential applications in low-resource language settings.

% In summary, our method provides a generalizable, scalable, and compute-efficient solution for augmenting vision-language models with continuous memory, enabling more effective reasoning over complex multimodal and multilingual inputs.
In summary, our proposed method, CoMEM, provides a generalizable, scalable, and compute-efficient solution for augmenting VLMs with a continuous memory mechanism. By leveraging the VLM itself as the memory encoder, CoMEM ensures strong semantic alignment between the memory and the model, while supporting seamless plug-and-play integration for diverse downstream tasks. This design enables effective reasoning over complex multimodal and multilingual inputs, offering a unified and efficient alternative to existing approaches that often rely on discrete context inputs, heavy fine-tuning, or multi-stage retrieval pipelines.

\section{Experiments}

\subsection{Experimental Setup}
\label{sec:experimental_setup}
\paragraph{Evaluation Settings}
We use WIT \citep{srinivasan2021wit} (Wikipedia-based Image Text Dataset) as our retrieval knowledge base. 
Building upon this, we conduct experiments across eight multimodal and multilingual reasoning benchmarks, including six multimodal reasoning benchmarks: InfoSeek \citep{chen2023can}, OVEN \citep{hu2023open}, MRAG-Bench \citep{hu2024mrag}, OK-VQA \citep{marino2019ok}, A-OKVQA \citep{schwenk2022okvqa}, and ViQuAE \citep{lerner2022viquae}, and two multilingual benchmarks: CVQA \citep{romero2024cvqa} and multilingual InfoSeek. Here we use GPT-4o-mini~\citep{openai2024gpt4o} to translate the InfoSeek from English into five different languages to match the language settings of CVQA.

Note (1) InfoSeek and OVEN are constructed from Wikipedia and consist of challenging factual questions.  
(2) MRAG-Bench, OK-VQA, A-OKVQA, and ViQuAE focus on multimodal real-world, knowledge-intensive tasks.   
(3) CVQA and multilingual InfoSeek evaluate model’s ability to reason diverse linguistic and cultural contexts. 
Further details about benchmarks are in Appendix \ref{appendix:benchmarks}.
% multilingual and multimodal reasoning, testing the model’s ability to generalize across languages and diverse linguistic contexts. Therefore, we can make sure that our evaluation is comprehensive and robust.

\paragraph{Baseline Methods}
We compare our method against three types of baselines: (1) VLMs, (2) VLMs with vanilla RAG, and (3) advanced RAG methods, covering a total of 18 different models.
%and methods to ensure a comprehensive and fair evaluation:

For VLMs, we evaluate their original capabilities on multimodal reasoning tasks without access to external knowledge, including: LLaVA-v1.5 \citep{liu2023visual}, LLaVA-v1.6 \citep{liu2023visual}, LLaVA-NeXT-LLaMA3 (denoted as LLaMA3 in tables) \citep{llava2024next}, InternLM-XComposer2.5vl (InternLM2.5vl) \citep{zhang2024internlm}, mPLUG-Owl3 \citep{ye2024mplugowl3}, Qwen2-VL-Instruct (Qwen2-VL) \citep{wang2024qwen2}, and Qwen2.5-VL-Instruct (Qwen2.5-VL) \citep{bai2025qwen2.5-vl}.

For VLMs with vanilla RAG, we directly insert retrieved image-text pairs into the input prompts of models, without making any architectural modifications or applying additional fine-tuning. This setup evaluates the effectiveness of naive retrieval-based augmentation.

% For advanced RAG methods, we include four state-of-the-art approaches designed for knowledge-intensive VQA tasks: Wiki-LLaVA \citep{caffagni2024wikillava}, RORA-VLM \citep{qi2024rora}, EchoSight \citep{echosight}, and ReflectiVA \citep{reflectiva}. Both Wiki-LLaVA and RORA-VLM adopt carefully designed two-stage retrieval frameworks to improve the relevance of retrieved knowledge. Building on this, ReflectiVA introduces reflective tokens that enable the VLM to filter relevant information through self-reflection. All three methods fine-tune the inference-time model itself. EchoSight trains a dedicated Q-Former module to perform more targeted retrieval based on the input query without inference-time model finetuning. Despite their effectiveness, all these methods ultimately rely on discrete context inputs, which can struggle with long or complex information.

Wiki-LLaVA and RORA-VLM use two-stage retrieval to improve knowledge relevance, while ReflectiVA adds reflective tokens for self-filtering. All three fine-tune the inference-time model. In contrast, EchoSight trains a separate Q-Former for retrieval without training the inference model. However, they all rely on discrete context inputs, which limits their ability to handle long contexts.

% Wiki-LLaVA and RORA-VLM adopt two-stage retrieval frameworks to improve the relevance of retrieved knowledge, while ReflectiVA extends this by introducing reflective tokens to help the VLM self-filter relevant information. These three methods fine-tune the model used at inference time. By contrast, EchoSight trains a separate Q-Former to retrieve targeted knowledge based on the input query without modifying the main model during inference. Despite their strengths, all rely on discrete context inputs, which can be limiting when handling long or complex knowledge contexts.

%In contrast, our method leverages a continuous memory mechanism to distill multimodal and multilingual information effectively. This allows us to achieve superior performance without requiring complex filtering strategies or extensive model modifications.

\paragraph{Implementation Details}
Our experimental pipeline comprises three phases: \textit{Knowledge Retrieval}, \textit{Knowledge Compression}, and \textit{Answer Generation}. To ensure fairness, we consistently use the top-10 retrieved image-text pairs across all experimental settings. We evaluate our method on Qwen2-Instruct-VL and Qwen2.5-Instruct-VL, demonstrating its strong generalization capability across different VLMs and question types. More implementation details can be found in Appendix \ref{appendix:details}.

\subsection{Main Results}

\paragraph{Evaluation on Multimodal Reasoning Task}
Table~\ref{tab:multimodal-eval} presents the performance comparison across six multimodal reasoning benchmarks, categorized into \emph{Base Models}, \emph{Retrieval-Augmented Baselines}, and our \emph{Continuous Memory} approach. Among base models, Qwen2-VL and Qwen2.5-VL achieve the highest performance across most benchmarks, which is likely due to their extensive multimodal training corpus and strong vision-language alignment. However, standard RAG integration often leads to inefficiencies in processing longer multimodal inputs, resulting in unstable performance that sometimes underperforms base models. To address this issue, advanced RAG models incorporate mechanisms that retrieve and use relevant content more effectively, resulting in improved performance on reasoning tasks. However, as shown in Section~\ref{subsec:comparison}, existing methods still face limitations, such as difficulty in adapting across modalities or a lack of generalizability across diverse task settings.

In comparison, our approach shows significant gains across multimodal reasoning benchmarks, with particularly strong improvements (over 15\%) on OKVQA and A-OKVQA versus baselines. These advancements originate from our VLM-based continuous memory architecture, which exhibits both strong adaptability to different VLMs and excellent generalization across diverse tasks. Remarkably, this level of performance requires minimal fine-tuning (1.2\% of parameters on 15.6k samples from InfoSeek, OKVQA and EVQA subsets), yet still achieves remarkable improvements on unseen benchmarks such as OVEN and A‑OKVQA. This suggests that our method can effectively fuse multimodal long-context knowledge, and generalize effectively to a wide range of downstream tasks.
\begin{table}[t]
\centering
\caption{Performance comparison with three types of baselines on knowledge-intensive VQA benchmarks. Bold indicates the best performance, and underscore denotes the second-best.}
\small
\resizebox{\textwidth}{!}{%
% \makebox[\textwidth][c]{%
\begin{tabular}{lccccccccccccc}
\toprule
 & \multicolumn{2}{c}{\textbf{InfoSeek}} & \multicolumn{2}{c}{\textbf{OVEN}} & \multirow{2}{*}{\textbf{MRAG}}& \multirow{2}{*}{\textbf{OKVQA}} & \multirow{2}{*}{\textbf{AOKVQA}} & \multirow{2}{*}{\textbf{ViQuAE}} &\multirow{2}{*}{\textbf{Avg.}}\\
\cmidrule(lr){2-3} \cmidrule(lr){4-5}
\textbf{Model} & \footnotesize Q & \footnotesize E & \footnotesize Q & E \\
\midrule
% \rowcolor{gray!20}
% \multicolumn{2}{l}{\textbf{Base-Models}}& & & & & & & &\\
LLaVA-v1.5  & 8.3 & 8.9 & 20.0 & 3.4 & 34.6 & 17.0 & 17.4 & 11.1 & 15.1\\
LLaVA-v1.6 & 10.3 & 9.1 & 17.9 & 1.8 & 33.4 & 31.4 & 31.7 & 18.7 & 19.3\\
LLaMA3 & 10.7 & 8.6 & 16.8 & 0.8 & 33.5 & 23.7 & 25.3 & 17.2 & 17.1\\
InternLM-2.5vl & 13.4 & 10.8 & 14.5 & 3.3 & 34.8 & 29.1& 32.8 & 29.7 & 19.5\\
mPLUG-Owl3 & 9.6 & 6.4 & 20.7&1.9&\textbf{45.0}&31.9&33.0&23.1 & 21.4\\
Qwen2-VL & 17.9 & 17.8 & 25.5 & 9.3 & 39.3 & 36.3 & 41.8 & 34.5 & 27.8\\
Qwen2.5-VL & 22.5 & 22.4 & 29.3 & 16.3 & 42.0 & 35.0 & 39.8 & \textbf{39.0} & 30.8\\
\midrule
%\rowcolor{gray!20}
%\multicolumn{4}{l}{\textbf{Retrieval-Augmented Models}}& & & & & & \\
LLaVA-v1.5 + RAG & 14.6 & 11.4 & 11.7 & 7.6 & 34.7 & 9.8 & 8.7 & 7.6 & 13.3\\
LLaVA-v1.6 + RAG& 6.7 & 5.8 & 9.7 & 1.2 & 32.6 & 25.6 & 22.6 & 17.0 & 15.2\\
LLaMA3 + RAG &12.1&	10.8&24.7	&\underline{21.5}&	36.4&20.7& 22.1&	18.1 & 20.8\\
InternLM-2.5vl + RAG &10.5&9.5&15.2&13.6&34.3&25.9&27.8&29.6 & 20.8\\
mPLUG-Owl3 + RAG &12.6&7.2&18.0&12.0&41.9&24.7&26.4&22.5 & 20.7\\
Qwen2-VL + RAG & 22.7 & 19.0 & 24.7 & \underline{21.5} & 40.4 & 41.9 & 45.3 & 33.6 & 31.1\\
Qwen2.5-VL + RAG & 17.7 & 18.8 & 23.0 & 19.7 & \underline{42.1} & 31.3 & 34.9 & 33.5 & 27.6\\
\hline
%\rowcolor{gray!20}
%\multicolumn{4}{l}{\textbf{Retrieval-Augmented Models with Spectial Architectures}}& & & & & &\\
Wiki-LLaVA & 28.6 & 25.7 & - & - & - & - & - & - & 27.2\\
RORA & 27.3 & 25.1 & 26.2 & 15.1 & - & - & - & - & 22.9\\
EchoSight & 18.0 & 19.8 & - & - & 41.3 & 20.0 & 16.9 & 25.2 & 23.5\\
ReflectiVA & 28.6 & 28.1  & - & - & 39.7 & 47.5 & 47.6 & 29.8 & \underline{36.9}\\
\midrule
\midrule
\rowcolor{cyan!8}
\textbf{CoMEM + Qwen2VL} & \underline{32.6} & \textbf{33.1} & \textbf{30.5} & \textbf{23.6} & 35.1 & \textbf{57.7} & \textbf{60.6} & \underline{36.3} & \textbf{38.7}\\
\rowcolor{cyan!8}
\textbf{CoMEM + Qwen2.5VL} & \textbf{32.8} & \underline{28.5} & \underline{26.0} & 20.8 & 38.1 & \underline{47.6} & \underline{55.0} & 34.7 & 35.4\\
\bottomrule
\end{tabular}%
}
% }
\label{tab:multimodal-eval}
\end{table}
% \end{center}

\paragraph{Evaluation on Multimodal Multilingual Reasoning Task}

We further evaluate our model's multilingual reasoning capabilities on the multilingual InfoSeek and CVQA benchmarks. As shown in Table~\ref{tab:model_comparison}, standard RAG methods demonstrate reduced effectiveness for non-English questions, potentially due to misalignment between retrieved multilingual content and input queries. In contrast, our memory mechanism encodes and stores transferable semantic representations that preserve core cross-modal and cross-lingual knowledge. This design translates into consistent accuracy improvements across all evaluated languages, achieving absolute gains of 6–12 points on InfoSeek-All scores while simultaneously showing enhanced performance on CVQA metrics. Notably, the model achieves particularly strong performance gains for Bulgarian (18\%) and Russian (10\%), underscoring the value of our language‑agnostic memory mechanism for lower‑resource settings where high‑quality retrieval is hardest to obtain. Overall, the results show that our method enables more robust grounding of multilingual queries and enhances reasoning capabilities across diverse tasks.

% This architecture enables more robust grounding of multilingual queries and enhances reasoning effectiveness across all evaluated benchmarks.

\begin{table}[h]
\caption{Performance comparison on Multilingual knowledge-intensive VQA benchmarks.}
\centering
% \small
\resizebox{\linewidth}{!}{ 
\begin{tabular}{llcccccccc}
\toprule
  \multirow{3}{*}{\textbf{Language}} & \multirow{3}{*}{\textbf{Method}} & \multicolumn{4}{c}{\textbf{Qwen2.5-Instruct-VL}} & \multicolumn{4}{c}{\textbf{Qwen2-Instruct-VL}} \\
\cmidrule(lr){3-6} \cmidrule(lr){7-10}
& & \multicolumn{3}{c}{\textbf{Multilingual InfoSeek}} & \multirow{2}{*}{\textbf{CVQA}} & \multicolumn{3}{c}{\textbf{Multilingual InfoSeek}} & \multirow{2}{*}{\textbf{CVQA}}\\
\cmidrule(lr){3-5} \cmidrule(lr){7-9}
 & & Unseen-Q & Unseen-E & All & & Unseen-Q & Unseen-E & All & \\
\midrule
\multirow{3}{*}{Chinese} & - & 17.4 & 13.8 & 15.4 & \textbf{82.32} & 15.1 & 10.9 & 12.6 & \textbf{74.60} \\
 & + RAG & 14.8 & 9.8 & 11.8 & 74.60 & 11.5 & 8.9 & 10.1 & 72.35 \\
 & + CoMEM & \textbf{22.5} & \textbf{21.5} & \textbf{22.0} & 78.46 & \textbf{23.1} & \textbf{19.5} & \textbf{21.1} & 73.31 \\
\midrule
\multirow{3}{*}{Russian} & - & 15.6 & 14.7 & 15.1 & 66.50 & 13.0 & 13.8 & 13.4 & \textbf{71.00} \\
 & + RAG & 10.6 & 8.9 & 9.7 & 66.00 & 13.6 & 10.9 & 12.1 & 62.50 \\
 & + CoMEM & \textbf{21.8} & \textbf{21.3} & \textbf{21.5} & \textbf{70.00} & \textbf{19.3} & \textbf{20.4} & \textbf{19.8} & \textbf{71.00} \\
\midrule
\multirow{3}{*}{Spanish} & - & 17.3 & 16.5 & 16.9 & 75.79 & 16.7 & 16.7 & 16.7 & 72.64 \\
 & + RAG & 12.3 & 11.4 & 11.8 & 79.25 & 9.5 & 8.1 & 8.7 & \textbf{76.10} \\
 & + CoMEM & \textbf{24.0} & \textbf{23.3} & \textbf{23.6} & \textbf{79.87} & \textbf{23.0} & \textbf{21.8} & \textbf{22.4} & 75.47 \\
\midrule
\multirow{3}{*}{Portuguese} & - & 18.7 & 18.1 & 18.4 & 66.55 & 18.4 & 19.3 & 18.8 & 66.90 \\
 & + RAG & 15.8 & 13.8 & 14.7 & 62.32 & 13.7 & 13.5 & 13.6 & \textbf{70.07} \\
 & + CoMEM & \textbf{27.1} & \textbf{27.2} & \textbf{27.2} & \textbf{66.90} & \textbf{24.1} & \textbf{26.1} & \textbf{25.1} & 67.96 \\
\midrule
\multirow{3}{*}{Bulgarian} & - & 12.5 & 12.0 & 12.2 & 46.09 & 8.0 & 7.9 & 7.9 & 45.55 \\
 & + RAG & 9.8 & 7.0 & 8.2 & 46.63 & 8.5 & 7.1 & 7.7 & 39.89 \\
 & + CoMEM & \textbf{19.3} & \textbf{17.4} & \textbf{18.3} & \textbf{47.44} & \textbf{15.9} & \textbf{18.3} & \textbf{17.0} & \textbf{50.13} \\
\midrule
\midrule
\multirow{3}{*}{Overall} & - & 17.3 & 16.2 & 16.7 & 67.45 & 14.8 & 14.4 & 14.6 & 66.14 \\
 & + RAG & 13.8 & 11.4 & 12.5 & 65.76 & 13.3 & 11.3 & 12.1 & 64.18 \\
 & + CoMEM & \textbf{24.9} & \textbf{23.6} & \textbf{24.2} & \textbf{68.53} & \textbf{23.0} & \textbf{23.2} & \textbf{23.1} & \textbf{67.57} \\
\bottomrule
\end{tabular}
}
\label{tab:model_comparison}
\end{table}

\subsection{Further Analysis}

\paragraph{Long Context Understanding Study}

To evaluate the ability of models to handle long-context inputs, we compare our method against vanilla RAG under varying numbers of retrieved image-text knowledge pairs. Specifically, we evaluate Qwen2-VL-Instruct and Qwen2.5-VL-Instruct on Infoseek, using both vanilla RAG and our method across different top-$k$ retrieval settings (from 3 to 50). 

\begin{wrapfigure}[13]{r}{0.56\textwidth}
\vspace{-15pt}
    \centering
    \includegraphics[width=0.45\textwidth]{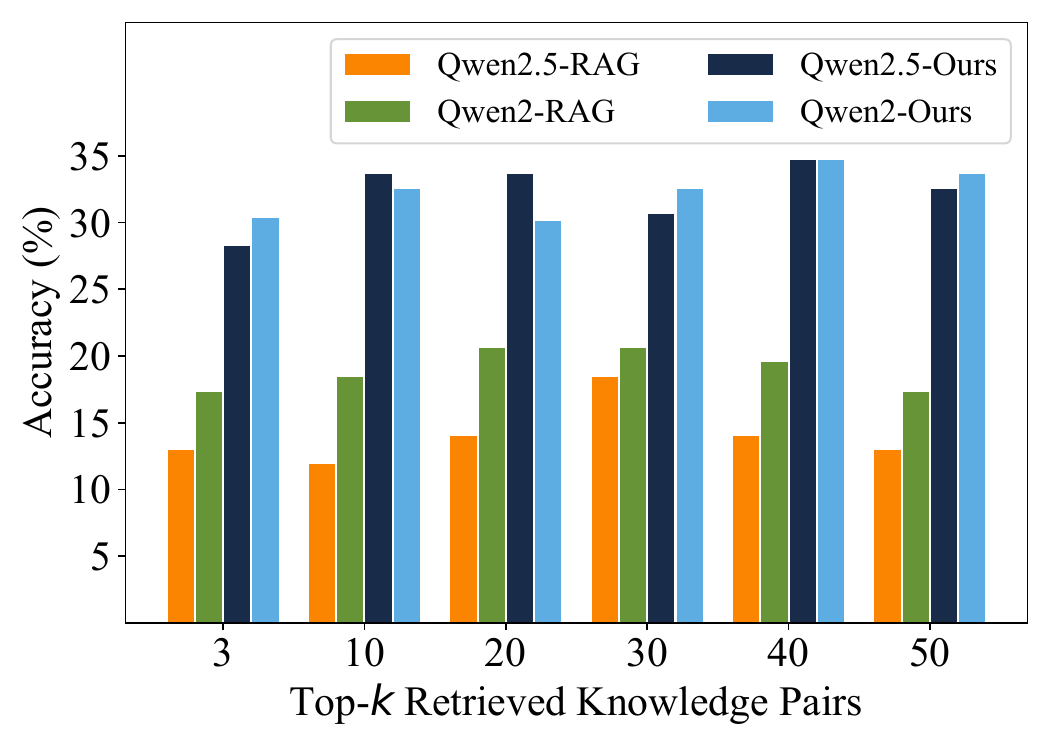}
    \caption{Comparison of Long Context Ability of RAG and Ours on Infoseek.}
    \label{fig:long_context}
\end{wrapfigure}

As shown in Figure \ref{fig:long_context}, the results reveal a clear trend: RAG-based performance begins to degrade when more than 30 retrieved pairs are added, but our method remains stable and performs consistently well across all retrieval sizes. These findings show that discrete token-based methods struggle with long context, while continuous memory enables scalable and reliable long-context reasoning. This robust performance as context length increases underscores the advantage of our approach in processing long, information-dense inputs. 

\begin{table}[h]
\centering
\small

\caption{Transferability Study of vision-language memory encoded by CoMEM on LLMs} %LMs
\resizebox{\textwidth}{!}{  % Resizes to text width while maintaining aspect ratio
% \small
\setlength{\tabcolsep}{6pt}
\renewcommand{\arraystretch}{1.2}
\begin{tabular}{lccccccc}
\toprule
 & \multicolumn{3}{c}{\textbf{InfoSeek(\%)}} & \multicolumn{3}{c}{\textbf{OVEN(\%)}} & \multicolumn{1}{c}{} \\
\cmidrule(lr){2-4} \cmidrule(lr){5-7}
\textbf{LLM}& \textbf{Unseen-Q} & \textbf{Unseen-E} & \textbf{All} & \textbf{Query} & \textbf{Entity} & \textbf{All} & \textbf{Avg.} \\
\midrule
Qwen2.5-Instruct & 5.0 & 4.8 &4.9 &2.4 &0.1 & 1.3& 3.1\\
Qwen2.5-Instruct + RAG & 13.4 & 10.3 &11.9 & 1.8& 2.7& 2.2& 7.0\\
Qwen2.5-Instruct + \textbf{CoMEM (using VLM)} &\textbf{29.3}&\textbf{27.4}&\textbf{28.3} &\textbf{6.8} & \textbf{7.7}&\textbf{7.2} &\textbf{17.8} \\
\bottomrule
\end{tabular}
}

\label{tab:transferability}
\end{table}

\paragraph{Transferability Study to LLMs.}
\label{sec:transfer}
To investigate whether the multimodal and multilingual continuous memory generated by a VLM can be effectively transferred to and leveraged by a pure Large Language Model (LLM), we conduct a transferability study. Specifically, we use Qwen2.5-VL-Instruct to encode visual and textual knowledge into dense continuous memory, and appended to the input embeddings of Qwen2.5-Instruct, a language-only LLM without vision capabilities. 

We evaluate our approach on InfoSeek and OVEN. As shown in Table \ref{tab:transferability}, our approach significantly outperforms both the vanilla LLM and the LLM augmented with text RAG, achieving an average accuracy of 17.8\%, compared to 7.0\% (RAG) and 3.1\% (baseline). These results demonstrate that LLMs can effectively leverage VLM-generated memory, even without vision modules.  This highlights a promising direction for cross-modal knowledge transfer, enabling LLMs to gain visual understanding through shared continuous memory without any architectural modifications.

% \paragraph{Training Efficiency Study}
% Training parameter number, data amount
% LoRA rank + Q-Former: 4, 2, 1, 1/2, 1/4
% Data Amount: 2, 1, 1/2, 1/4, 1/8
% Token: 2,4,16

\section{Related Work}
\paragraph{Vision-Language Models.}

LLMs have seen significant advancements, with models like GPT-4~\citep{openai2024gpt4o} and Qwen-2.5~\citep{yang2024qwen2.5} demonstrating emergent capabilities such as in-context learning and complex reasoning. Building upon these advancements, VLMs have emerged to integrate visual and textual modalities, enabling models to process and understand multimodal data. To effectively extend language understanding into the visual domain, VLMs combine specialized neural network architectures for vision processing (such as Vision Transformers) with language models, enabling joint reasoning over visual and textual inputs. These models are typically trained on large-scale datasets that pair images with descriptive text to learn joint representations, using techniques like contrastive learning~\citep{chen2020simple, radford2021learning}, multimodal pretraining~\citep{mckinzie2024mm1, xue2024ulip}, and instruction-aware tuning~\citep{liu2023visual, liu2024improved,liu2024less}.

% A key trend has been the scaling law of model sizes, leading to the emergence of models with tens an hundreds of billions of parameters, such as the GPT series\citep{radford2019language, brown2020language, achiam2023gpt}, Llama\citep{touvron2023llama} and Qwen. This scaling has unlocked emergent abilities in LLMs, including in-context learning, instruction following, and complex reasoning\citep{wei2022emergent}. 
% Many progress has also been made in training techniques to enhance the performance and efficiency of LLMs. These include methodologies like instruction tuning\citep{wei2021finetuned} and Reinforcement Learning from Human Feedback (RLHF)\citep{ouyang2022training} to better align models with human preferences and improve their ability to follow instructions\citep{zhao2023survey}. Furthermore, techniques such as low-precision training\citep{hubara2018quantized} and parameter-efficient fine-tuning (PEFT)\citep{hu2022lora} have been developed to reduce the computational resources required for training and adapting large language models.The remarkable capabilities of recent LLMs lave led to their widespread adoption across numerous domains. They are being explored and implemented in scientific research, code generation, content creation, and various specialized industries like healthcare and finance.

\paragraph{Context Compression.}

% A key challenge in unlocking the full potential of language models is their constrained context windows, which limits the amount of information they can process at once. To address this issue, context compression methods have been introduced to enable models to handle longer input sequences. These approaches can be categorized as training-free Gisting~\citep{mu2023learning} proposes a method that trains language models to compress prompts into a small set of reusable "gist tokens" by modifying Transformer attention masks. IC-Former~\citep{wang2024context} introduced a cross-attention-based lightweight model that condenses length inputs into compact digest vectors, achieving up to 112x speedups while preserving over 89\% of the original contextual information. Such techniques not only alleviate memory and latency constraints but also enhance the model's ability to perform complex reasoning tasks that require integrating extensive information.

The constrained context windows of language models limit their information processing capacity, prompting the development of context compression methods to enable longer-sequence handling. One of the approach towards context compression in LLMs is through token pruning. FastV\citep{chen2024image} distills vision-language knowledge into compact key-value memory slots, while SparseVLM\citep{zhang2024sparsevlm} selects a sparse subset of visual tokens via top-down routing. In contrast, Gisting\citep{mu2023learning} compress long prompts into a small set of reusable "gist tokens" by modifying Transformer attention masks. Another approach involves soft prompts, which introduce trainable vector embeddings to input sequences, enabling efficient task adaptation. IC-Former\citep{wang2024context} compresses long input sequences into compact digest vectors, while SPC-LLM\citep{wang2024adapting} combines natural language summarization with trainable soft prompts. Both methods condense lengthy input sequences into shorter representations, enhance the efficiency of LLMs and preserve over 90\% of the original performance.
%\subsection{Memory for Language Models}
% P-Tuning\citep{liu2021p} shows a small number of learned embeddings can carry rich task-relevant signals, while
\paragraph{Memory for Language Models.}

As LMs face limitations in context length and long-term information retention, memory mechanisms have emerged to enhance their capacity for information-intensive reasoning and knowledge storage. Early retrieval-based approaches such as RAG~\cite{lewis2020rag} and REALM~\cite{guu2020realm} retrieve external documents and inject them as long token sequences during inference time. However, these methods are constrained by context length limits and the inefficiency of discrete token representations, especially for supporting multimodal information.
Recent advances shift toward continuous memory, representing knowledge as dense vectors rather than raw text. Approaches like VoCo-LLaMA~\cite{vocolama} and MA-LMM~\cite{he2024malmm} compress visual content into compact embeddings. Concurrently, strategies for memory storage have evolved. Persistent memory systems such as LONGMEM~\cite{wang2023longmem} store compressed knowledge in cache key-value (KV) formats, while retrieval-based methods like WikiLLaVA~\cite{caffagni2024wikillava}, RORA-VLM~\cite{roravlm}, and EchoSight~\cite{echosight} treat external knowledge bases as memory banks, using dedicated retrieval frameworks to support VQA tasks.

\section{Conclusion}
In this paper, we empirically demonstrate that a VLM can effectively serve as its own memory encoder, capable of converting multimodal knowledge into compact continuous embeddings. Building on this insight, we develop a data- and parameter-efficient method to fine-tune the VLM as a continuous memory encoder. Specifically, by updating only 1.2\% of the model’s parameters using just 15.6k self-synthesized samples, the resulting memory module can encode diverse multimodal and multilingual knowledge into merely 8 continuous embeddings. Importantly, since the VLM remains unchanged during inference, our memory module can be seamlessly integrated or detached as needed. Extensive evaluations across six English and two multilingual vision-reasoning benchmarks demonstrate the effectiveness and versatility of our approach.
% In this paper, we empirically demonstrate that a VLM can serve as its own memory encoder, which can convert the multimodal knowledge into continuous embeddings.
% Then, we proposed a data-efficient and parameter-efficient approach to fine-tune the VLM into a continuous memory encoder. 
%Concretely, we only fine-tuned 1.2\% of its parameters on only 15.6k self-synthesized samples, and then the memory can effectively encode any multimodal and multilingual knowledge into only 8 continuous embeddings.Since we do not modify the inference-time VLM, our memory can be flexibly integrated with the VLM when necessary. Extensive experiments on six English and two multilingual vision-reasoning benchmarks, demonstrate the superiority of our approach.

In future work, we plan to extend our approach to a wider range of complex reasoning and planning tasks. Additionally, we aim to integrate the continuous memory mechanism into multimodal agents and evaluate its effectiveness in facilitating knowledge transfer across multiple language and vision-language models.
%In the future, we will explore broader applications of our approach on various complex reasoning and planning tasks. Besides, we will integrate our continuous memory mechanism into multimodal agents, and test its effectiveness in transmitting the knowledge among multiple LMs and VLMs.
%dynamic memory growth during dialogue, application to video-centric tasks, and broader transfer to purely textual LLMs.

\bibliographystyle{unsrt}
\bibliography{ref.bib}

%%%%%%%%%%%%%%%%%%%%%%%%%%%%%%%%%%%%%%%%%%%%%%%%%%%%%%%%%%%%

%%%%%%%%%%%%%%%%%%%%%%%%%%%%%%%%%%%%%%%%%%%%%%%%%%%%%%%%%%%%
\newpage
\appendix

% \section{Technical Appendices and Supplementary Material}
% Technical appendices with additional results, figures, graphs and proofs may be submitted with the paper submission before the full submission deadline (see above), or as a separate PDF in the ZIP file below before the supplementary material deadline. There is no page limit for the technical appendices.

\section{Benchmark Details}
\label{appendix:benchmarks}
\paragraph{InfoSeek} 
InfoSeek is a visual question answering (VQA) dataset tailored for information-seeking questions that cannot be answered with only common sense knowledge. It combines human-annotated and automatically collected data from visual entity recognition datasets and Wikidata, providing over one million examples for model fine-tuning and validation~\cite{chen2023can}.
For InfoSeek, the ground truth answers for test sets are not publicly available, so we follow prior work \citep{caffagni2024wikillava, echosight, reflectiva} and report results on the validation sets. These sets include questions not seen during training and those associated with unseen entities. 

\paragraph{OVEN} 
OVEN (Open-domain Visual Entity Recognition) challenges models to select among six million possible Wikipedia entities, making it a general visual recognition benchmark with the largest number of labels. It is constructed by re-purposing 14 existing datasets with all labels grounded onto one single label space: Wikipedia entities~\cite{hu2023open}.
Similar with Infoseek, the ground truth answers for the test sets of OVEN are not publicly available, so we also report results on the validation sets. 

\paragraph{MRAG-Bench} 
MRAG-Bench is a multimodal retrieval-augmented generation benchmark designed to evaluate the performance of large vision-language models (LVLMs) in scenarios where visual knowledge retrieval is more beneficial than textual information. It consists of 16,130 images and 1,353 human-annotated multiple-choice questions across nine distinct scenarios~\cite{hu2024mrag}.

\paragraph{OK-VQA} 
OK-VQA includes more than 14,000 open-ended questions that require external knowledge to answer. The dataset is manually filtered to ensure all questions necessitate information beyond the image content, such as from Wikipedia~\cite{marino2019ok}.

\paragraph{A-OKVQA} 
A-OKVQA is a crowdsourced visual question answering dataset composed of approximately 25,000 questions requiring a broad base of commonsense and world knowledge to answer. Unlike existing knowledge-based VQA datasets, the questions generally cannot be answered by simply querying a knowledge base and instead require some form of commonsense reasoning about the scene depicted in the image~\cite{schwenk2022okvqa}.

\paragraph{ViQuAE} 
ViQuAE is a dataset focusing on knowledge-based visual question answering about named entities. It covers a wide range of entity types, such as persons, landmarks, and products, and evaluates models' abilities to ground visual content with knowledge base information~\cite{lerner2022viquae}.

\paragraph{CVQA}
CVQA (Culturally-diverse Multilingual Visual Question Answering) dataset is a benchmark that offers a broad, inclusive representation by incorporating culturally-driven images and questions from a wide range of countries and languages\citep{romero2024cvqa}. In this study, we evaluate five of the most widely used languages in CVQA: Chinese, Russian, Spanish, Portuguese, and Bulgarian.

For all benchmarks, we follow the official evaluation protocols to compute the accuracy of the model's responses. Specifically:
(1) For InfoSeek, OK-VQA, A-OKVQA, and ViQuAE, we use exact match evaluation to verify whether the model’s response exactly matches the ground-truth answers.
(2) For OVEN, we adopt the official evaluation script, which uses BM25 \citep{robertson2009bm25} to match the model’s answer with relevant Wikipedia entities.
(3) For MRAG-Bench and CVQA, which are in multiple-choice format, we evaluate accuracy by checking whether the model selects the correct option.

\section{Implementation Details}
\label{appendix:details}

$\bullet$  \textbf{Knowledge Retrieval}
    Our knowledge base is constructed using the Wikipedia-based Image-Text (WIT) dataset\citep{srinivasan2021wit}, which consists of 37.5 million curated image-text pairs from Wikipedia articles across 108 languages. Based on WIT knowledge base, we implement a CLIP-based image-to-image retrieval system to identify the most relevant external knowledge. Following the stage-1 retrieval methodology of RoRA\citep{roravlm}, we first encode all images in WIT using a frozen CLIP image encoder\citep{radford2021learning} to build a dense vector-search database. Given a query image $\mathcal{I}$, its CLIP embedding $CLIP(\mathcal{I})$ is compared against all vectors in the knowledge base via cosine similarity, followed by softmax normalization over the similarity scores. The image retriever then returns the top-$k$ highest-scoring images along with their associated textual descriptions.
    
$\bullet$  \textbf{Memory Encoding} Given the retrieved image-text pairs, we employ a memory encoder, consisting of a VLM and a Q-Former to compress multimodal information. For Each image-text pair is compressed into an 8-token vector. These token vectors are then concatenated and passed into the inference-time model. For Qwen2.5-Instruct VL, we uses Qwen2.5-Instruct VL as both the inference-time model and the memory encoder, and for Qwen2-Instruct VL, we uses Qwen2-Instruct VL as both the inference-time model and the memory encoder. 

$\bullet$  \textbf{Answer Generation} The concatenated compressed tokens are plug into the inference-time model to generate answers. We should note that \textit{our compression module is model-agnostic, allowing the memory encoder to be plugged into other LMs.} This flexibility is further demonstrated in Section \ref{sec:transfer}.

\section{Ablation Study on Embedding Size}
\begin{table}[h]
%\vspace{-15pt}
\centering
\small
\caption{Performance on Infoseek of Different Embedding Sizes}
\label{tab:embedding-performance}
\vspace{5pt}
%\scalebox{0.75}{ % 缩放整个表格
%\begin{minipage}{1.33\linewidth} % 调整minipage宽度以补偿缩放
\begin{tabular}{lcccc}
\toprule
\textbf{Metric} & \textbf{4-Emb} & \textbf{8-Emb} & \textbf{16-Emb} & \textbf{24-Emb} \\
\midrule
Unseen Question Score & 29.32\% & \textbf{32.80\%} & 31.95\% & 31.37\% \\
Unseen Entity Score & 29.64\% & \textbf{32.33\%} & 30.03\% & 30.83\% \\
\midrule
Final Score & 29.48\% & \textbf{32.74\%} & 30.96\% & 31.10\% \\
\bottomrule
\end{tabular}
%\end{minipage}
%}
\end{table}

To identify the optimal memory size for CoMEM, we conduct an ablation on the number of continuous embeddings (4, 8, 16, 24) using the InfoSeek \citep{infoseek} benchmark. As shown in Table~\ref{tab:embedding-performance}, using 8 embeddings yields the best overall performance (32.74\%), outperforming both smaller and larger configurations. Fewer embeddings underrepresent complex multimodal knowledge, while more embeddings introduce redundancy and noise. This result confirms that 8 embeddings offer an effective trade-off between expressiveness and efficiency, supporting our design choice in CoMEM.

\section{Method Generalizability}

\subsection{Evaluation on Image Captioning}

we evaluate the generalizability of CoMEM on a caption generation task using the COCO 2014 dataset \citep{lin2014coco}. We randomly sampled 100 image-caption pairs from the test set and used CLIP to retrieve relevant image-caption pairs from the training set. We then compared three setups: (1) Original model (no retrieval); (2) RAG-style retrieval (retrieved image-caption pairs prepended in prompt); (3) Our method (retrieved image-caption pairs encoded into continuous memory). We tested both Qwen2.5 and Qwen2, and report standard captioning metrics below in Table \ref{tab:coco-captioning}. We obtain the following conclusions:

\paragraph{Better precision and fluency:} 
BLEU-1 / BLEU-4 measure n-gram precision; METEOR balances precision and recall with synonym matching; higher scores indicate better word accuracy. CoMEM consistently improves BLEU-1, BLEU-4, and METEOR across both models, indicating more fluent and accurate captions.
\paragraph{Substantial gain in content relevance:} CIDEr scores use tf-idf weighted n-grams to measure relevance to ground-truth captions. It's increased significantly in Qwen2.5-VL with CoMEM (from 0.24 → 0.64), showing better content relevance to ground-truth captions.
\paragraph{Improved semantic similarity:} ROUGE-L captures the longest common subsequence with the reference, reflecting surface-level overlap. BERTScore-F computes semantic similarity using contextual embeddings. Memory-augmented models generate captions with high ROUGE-L and BERTScore-F scores, showing stronger lexical overlap and semantic similarity.

These results confirm that CoMEM generalizes well beyond QA tasks, offering an effective and lightweight memory mechanism for open-ended generation tasks like image captioning.

\begin{table}[h]
\centering
\small
\caption{Evaluation Metrics for Qwen2.5-VL and Qwen2-VL on COCO Captioning task}
\label{tab:coco-captioning}
\vspace{0.5em}
\resizebox{\textwidth}{!}{
\begin{tabular}{llccccccc}
\toprule
\textbf{Model} & \textbf{Method} & \textbf{BLEU-1} & \textbf{BLEU-4} & \textbf{METEOR} & \textbf{CIDEr} & \textbf{ROUGE-L} & \textbf{BERTScore-F} & \textbf{Trigram Diversity} \\
\midrule
\multirow{3}{*}{qwen2.5-VL} 
& Original  & 0.26 & 0.04 & 0.18 & 0.24 & 0.26 & 0.81 & 0.92 \\
& +RAG      & 0.30 & 0.05 & 0.20 & 0.45 & 0.27 & 0.81 & \textbf{0.94} \\
& +CoMEM    & \textbf{0.34} & \textbf{0.07} & \textbf{0.21} & \textbf{0.64} & \textbf{0.32} & 0.81 & 0.86 \\
\midrule
\multirow{3}{*}{qwen2-VL} 
& Original  & 0.34 & 0.08 & 0.21 & 0.79 & 0.34 & 0.81 & 0.84 \\
& +RAG      & 0.34 & 0.07 & 0.20 & 0.64 & 0.32 & 0.81 & 0.83 \\
& +CoMEM    & \textbf{0.36} & 0.08 & \textbf{0.23} & 0.76 & 0.32 & \textbf{0.82} & \textbf{0.88} \\
\bottomrule
\end{tabular}
}
\end{table}

\subsection{Evaluation on Reasoning Tasks}

\begin{table}[h]
\centering
\small
\caption{Performance Comparison on Reasoning Task MMMU}
\label{tab:mmmu}
\vspace{0.5em}
\resizebox{\textwidth}{!}{
\begin{tabular}{llccccccccc}
\toprule
\textbf{Metrics} & \textbf{Method} & \textbf{Accounting} & \textbf{Architecture} & \textbf{Clinical Med} & \textbf{Computer} & \textbf{Economics} & \textbf{Electronics} & \textbf{Management} & \textbf{Materials} & \textbf{Pharmacy} \\
\midrule
\multirow{3}{*}{Accuracy} 
& Qwen2.5-VL     & 26.7 & 23.3 & 43.3 & 26.7 & 46.7 & 23.3 & \textbf{40.0} & 30.0 & 30.0 \\
& +RAG           & 23.3 & 26.7 & 40.0 & 33.3 & 33.3 & 16.7 & 36.7 & 33.3 & \textbf{40.0} \\
& +CoMEM         & \textbf{40.0} & 26.7 & \textbf{43.3} & \textbf{40.0} & \textbf{46.7} & \textbf{30.0} & 36.7 & \textbf{33.3} & \textbf{40.0} \\
\midrule
\multirow{3}{*}{Similarity} 
& Qwen2.5-VL     & 26.7 & 26.7 & 43.3 & 26.7 & 46.7 & 26.7 & \textbf{40.0} & 30.0 & 30.0 \\
& +RAG           & 23.3 & 26.7 & 40.0 & 33.3 & 33.3 & 16.7 & 36.7 & 33.3 & \textbf{40.0} \\
& +CoMEM         & \textbf{40.0} & \textbf{33.3} & \textbf{43.3} & \textbf{40.0} & \textbf{50.0} & \textbf{30.0} & \textbf{43.3} & \textbf{33.3} & \textbf{43.3} \\
\bottomrule
\end{tabular}
}
\end{table}

\begin{wraptable}{r}{0.55\textwidth}
\centering
\small
\caption{Performance Comparison on Reasoning Task MathVista}
\label{tab:mathvista}
\vspace{0.5em}
\resizebox{0.55\textwidth}{!}{
\begin{tabular}{llcccc}
\toprule
\textbf{Metrics} & \textbf{Method} & \textbf{Math} & \textbf{Textbook QA} & \textbf{Visual QA} & \textbf{Figure QA} \\
\midrule
\multirow{3}{*}{Accuracy} 
& Qwen2.5-VL     & 59 & 50 & 34 & 46 \\
& +RAG           & 63 & \textbf{53} & 37 & 54 \\
& +CoMEM         & \textbf{65} & 46 & \textbf{39} & \textbf{56} \\
\midrule
\multirow{3}{*}{Similarity} 
& Qwen2.5-VL     & 62 & 55 & 35 & 48 \\
& +RAG           & 63 & \textbf{57} & 37 & 56 \\
& +CoMEM         & \textbf{66} & 49 & \textbf{41} & \textbf{57} \\
\bottomrule
\end{tabular}
}
\end{wraptable}

In recent LLM/MLLM reasoning studies, various reasoning-heavy benchmarks such as MathVista \citep{lu2023mathvista} and MMMU \citep{yue2023mmmu} have been widely adopted. It would be interesting to see whether the proposed method can effectively handle reasoning tasks on these benchmarks.

Concretely, we split the dataset into evaluation set and retrieval set, and retrieve relevant image-text pairs using CLIP in RAG and Memory settings. We calculate accuracy with exact matching and similarity with overlapping ratio between predicted and ground truth tokens.

Our results show that CoMEM consistently improves performance across diverse domains:

On MMMU (Table \ref{tab:mmmu}), our method consistently outperforms both the Original and RAG baselines across most domains, especially in domains like Accounting, Computer Science, Pharmacy, Electronics, etc.
On MathVista (Table \ref{tab:mathvista}), we also observe improvements in different types of reasoning-heavy tasks. These results demonstrate that our proposed method effectively enhances accuracy in complex reasoning scenarios across domains, showing the robustness and generalizability of our method.

\section{Method Cost}
\subsection{Inference Latency Test}

To test the latency of baselines and our method, we evaluated the throughput (tokens per second) for each model on 100 random sampled instances. The higher throughput indicates faster inference speed. All experiments were conducted on a single NVIDIA H100 and the results are summarized in Table \ref{tab:latency}.

The findings show that CoMEM maintains competitive inference speed as the original model, because it does not significantly increase input length. For RAG, as it expands the input by up to 15× and causes a proportional increase in attention cost, the inference speed decreases a lot. Since our CoMEM adds less than 100 continuous memory tokens, this keeps inference cost close to that of the base model while improving accuracy.

\begin{table}[h]
\centering
\small
\caption{Inference Latency Evaluation accross different benchmarks }
\label{tab:latency}
\vspace{0.5em}
\begin{tabular}{lcccccc}
\toprule
\textbf{Token per Second} & \textbf{InfoSeek} & \textbf{OVEN} & \textbf{MRAG} & \textbf{OKVQA} & \textbf{AOKVQA} & \textbf{ViQuAE} \\
\midrule
\rowcolor{gray!10} \multicolumn{7}{l}{\textbf{Qwen2 models}} \\
Qwen2-VL            & 55.93 & 100.91 & 39.54 & 45.81 & 53.38 & 99.53 \\
Qwen2-VL+RAG        & 40.71 & 49.27  & 8.38  & 34.66 & 23.60 & 37.50 \\
Qwen2-VL+COMEM      & 48.93 & 53.29  & 18.61 & 40.42 & 51.60 & 41.50 \\
\addlinespace
\rowcolor{gray!10} \multicolumn{7}{l}{\textbf{Qwen2.5 models}} \\
Qwen2.5-VL          & 53.44 & 72.71  & 42.13 & 45.64 & 42.87 & 53.49 \\
Qwen2.5-VL+RAG      & 41.17 & 42.04  & 15.89 & 32.22 & 27.93 & 39.61 \\
Qwen2.5-VL+CoMEM    & 51.05 & 54.39  & 19.79 & 42.86 & 61.41 & 48.81 \\
\bottomrule
\end{tabular}
\end{table}

\subsection{Token Cost Comparison}

\begin{table}[h]
\centering
\small
\caption{Token Number and Accuracy Across Datasets}
\label{tab:token-accuracy}
\vspace{0.5em}
\resizebox{\textwidth}{!}{  % Resize to page width and auto height
\begin{tabular}{lcccccc}
\toprule
\textbf{Token Number (Accuracy\%)} & \textbf{Infoseek} & \textbf{Oven} & \textbf{MRAG} & \textbf{OKVQA} & \textbf{AOKVQA} & \textbf{ViQuAE} \\
\midrule
\textbf{Qwen2-VL} & 357.6 (17.8) & 408.6 (25.5) & 84.5 (39.3) & 404.8 (36.3) & 418.4 (41.8) & 424.7 (34.5) \\
\textbf{Qwen2-VL+RAG} & 6163.8 (19.0) & 6481.5 (24.7) & 5882.4 (40.4) & 6803.8 (41.9) & 6715.4 (45.3) & 6943.2 (33.6) \\
\textbf{Qwen2-VL+COMEM} & 416.8 (33.1) & 492.9 (30.5) & 176.9 (35.1) & 453.8 (57.7) & 478.3 (60.6) & 489.2 (36.3) \\
\bottomrule
\end{tabular}
} % end resizebox
\end{table}

Continuous embeddings better support high compression, as continuous embeddings can densely encode the information, which prevents the great increase of context length (see Table\ref{tab:token-accuracy}). It is rather helpful for VLMs to read and understand massive multimodal data. For the example of RAG scenario, here we list the context length in three settings: Original model(no retrieval) , Qwen2-VL+RAG and our CoMEM method. We can clearly see that although we only retrieve 10 examples in the context, the context length increases by approximately 15× when using RAG. It will cause the VLM to suffer from the long context understanding problem. In contrast, our method converts the retrieved examples into continuous embeddings, which leads to fewer than 100 additional embeddings to the input sequence. It avoids the long context understanding problem and also reduces the cost of encoding such long input sequences.

\section{Training Efficiency}

\begin{wraptable}[17]{r}{0.58\textwidth}
\vspace{-10pt}
\centering
\begin{tabular}{llccc}
\toprule
& & \multicolumn{3}{c}{\textbf{Infoseek}}\\
\cmidrule(lr){3-5}
\multicolumn{2}{l}{\textbf{Training Settings}} & Unseen-Q & Unseen-E  & All \\
\midrule
Original & & 32.8 & 28.5 & 30.7 \\
\midrule
\midrule
\multirow{4}{*}{Data} 
 & 4x   & \textbf{34.8} & 28.4 & \textbf{31.3} \\
 & 2x   & 32.2 & \textbf{29.8} & 30.9 \\
 & 0.5x & 26.5 & 24.4 & 25.4 \\
 & 0.25x & 17.8 & 17.5 & 17.6 \\
\midrule
\multirow{4}{*}{Parameters} 
 & 4x   & 26.4 & 22.1 & 24.1 \\
 & 2x   & 28.6  & 24.8 & 26.3 \\
 & 0.5x & 27.8  & 24.7 & 26.1 \\
 & 0.25x & 23.1 & 20.3 & 21.6 \\
\bottomrule
\end{tabular}
\caption{Performance of CoMEM on Qwen2.5-VL under different training data and parameter settings.}
\label{tab:efficiency}
\end{wraptable}

To evaluate the training efficiency of our method, we assess the performance of CoMEM on Qwen2.5-VL using the Infoseek benchmark under varying amounts of training data and trainable parameters.
In the original setting, we use only 15.6k training samples and fine-tune 1.2\% of the total parameters. For the data variation setting, we scale the training data by factors of 0.25×, 0.5×, 2×, and 4×. For the parameter variation setting, we adjust the LoRA rank and the number of Q-Former layers by the same scaling factors to control the number of trainable parameters.

As shown in Table~\ref{tab:efficiency}, increasing the training data by 2× or even 4× results in only marginal performance gains, suggesting that the original data size is already adequate for effective training. Similarly, increasing the number of trainable parameters does not yield improvements, while reducing them below the original configuration leads to a notable drop in performance. These findings highlight that our training recipe is both data- and parameter-efficient, achieving strong results with minimal resource expenditure.

\section{Limitations}
\label{appendix:limitations}

$\bullet$  \textbf{Evaluation Benchmarks} 
While we evaluate our method on 6 multimodal and 2 multilingual reasoning tasks, most of benchmarks are static and synthetic. Real-world applications with dynamic or noisy inputs (\eg web data, live video) may introduce challenges.

$\bullet$ \textbf{Multi-Agent Settings} 
Our current framework is designed and evaluated in a single-model setting, where one inference language model uses the continuous memory module for enhanced reasoning. However, many real-world applications involve multiple collaborating agents or a combination of LMs and VLMs.  Whether our continuous memory can effectively transmit and share knowledge across multiple models remains unexplored and will be investigated in future work.

\section{Case Study}

\begin{figure}[h]
    \centering
    \includegraphics[width=1\linewidth]{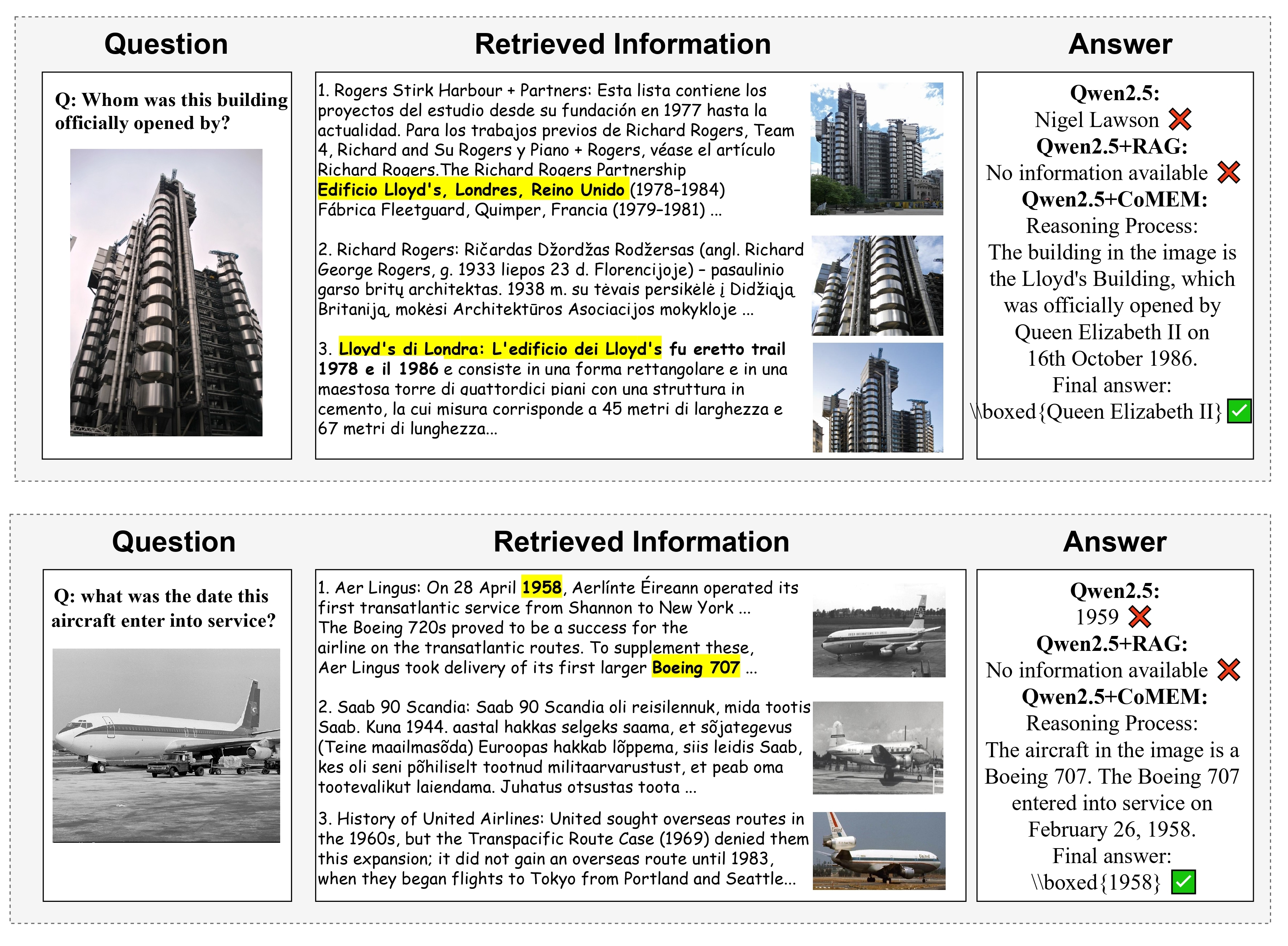}
    \caption{Case studies comparing CoMEM with baseline model and model with RAG.}
    \label{fig:enter-label}
\end{figure}

In this appendix, we present a qualitative case study to demonstrate the effectiveness of our proposed model. Given a question and a corresponding query image, our pipeline first retrieves the top 10 relevant image-text pairs from the WIT knowledge base to provide rich contextual information. Due to space constraints, we only display three representative retrieved pairs for each example in this appendix. We then compare the performance of our CoMEM model against two baselines: the standalone Qwen2.5-VL and a baseline retrieval-augmented generation (RAG) model. CoMEM can effectively capture key information from retrieved supporting texts, even when the exact answer is not explicitly provided, and perform reasoning to derive the correct answer.

These case studies demonstrate that CoMEM is able to generate accurate answers in challenging scenarios where baseline models either fail or return incomplete information. This highlights CoMEM’s ability to effectively encode and leverage complex multimodal and multilingual knowledge, leading to stronger performance in advanced reasoning tasks.

% \begin{figure}[htbp]
%     \centering
%     \includegraphics[width=1\linewidth]{plots/case1.png}\\
%     \includegraphics[width=1\linewidth]{plots/case2.png}\\
%     \includegraphics[width=1\linewidth]{plots/case3.png}
    
%     \label{fig:case}
% \end{figure}

\newpage
% \ignore{
\section*{NeurIPS Paper Checklist}

\begin{enumerate}

\item {\bf Claims}
    \item[] Question: Do the main claims made in the abstract and introduction accurately reflect the paper's contributions and scope?
    \item[] Answer: \answerYes % Replace by \answerYes{}, \answerNo{}, or \answerNA{}.
    \item[] Justification: The abstract and introduction clearly and accurately reflect the paper's main contributions and scope. The proposed method uses VLM itself as a continuous memory encoder with minimal fine-tuning effort, which addresses the issues exist in existing RAG-based and token-pruning models. Our approach is validated by the result of massive experiment, which is shown in ~\ref{tab:multimodal-eval}. 

\item {\bf Limitations}
    \item[] Question: Does the paper discuss the limitations of the work performed by the authors?
    \item[] Answer: \answerYes{} % Replace by \answerYes{}, \answerNo{}, or \answerNA{}.
    \item[] Justification: We discuss the limitations of the work in Appendix~\ref{appendix:limitations}

\item {\bf Theory assumptions and proofs}
    \item[] Question: For each theoretical result, does the paper provide the full set of assumptions and a complete (and correct) proof?
    \item[] Answer: \answerNA{} % Replace by \answerYes{}, \answerNo{}, or \answerNA{}.
    \item[] Justification: We do not include theoretical results in this paper.

    \item {\bf Experimental result reproducibility}
    \item[] Question: Does the paper fully disclose all the information needed to reproduce the main experimental results of the paper to the extent that it affects the main claims and/or conclusions of the paper (regardless of whether the code and data are provided or not)?
    \item[] Answer: \answerYes{} % Replace by \answerYes{}, \answerNo{}, or \answerNA{}.
    \item[] Justification: We disclose all of our experiment details in ~\ref{sec:experimental_setup} and Appendix~\ref{appendix:details}. The information provided in these two sections is enough for reproducing the results. All of our code and data will be released. 

\item {\bf Open access to data and code}
    \item[] Question: Does the paper provide open access to the data and code, with sufficient instructions to faithfully reproduce the main experimental results, as described in supplemental material?
    \item[] Answer: \answerYes{} % Replace by \answerYes{}, \answerNo{}, or \answerNA{}.
    \item[] Justification: We will release our code, data, and model checkpoints for review, and open-source them upon paper acceptance.

\item {\bf Experimental setting/details}
    \item[] Question: Does the paper specify all the training and test details (e.g., data splits, hyperparameters, how they were chosen, type of optimizer, etc.) necessary to understand the results?
    \item[] Answer: \answerYes{} % Replace by \answerYes{}, \answerNo{}, or \answerNA{}.
    \item[] Justification: We disclose all of our experiment details in ~\ref{sec:experimental_setup} and Appendix~\ref{appendix:details}. The information provided in these two sections is enough for reproducing the results. All of our code and data will be released.

\item {\bf Experiment statistical significance}
    \item[] Question: Does the paper report error bars suitably and correctly defined or other appropriate information about the statistical significance of the experiments?
    \item[] Answer: \answerNo{} % Replace by \answerYes{}, \answerNo{}, or \answerNA{}.
    \item[] Justification: We do not include statistical significance results, as we have achieved apparent and great improvements, and there is no need for significance test.

\item {\bf Experiments compute resources}
    \item[] Question: For each experiment, does the paper provide sufficient information on the computer resources (type of compute workers, memory, time of execution) needed to reproduce the experiments?
    \item[] Answer: \answerYes{} % Replace by \answerYes{}, \answerNo{}, or \answerNA{}.
    \item[] Justification: We clearly disclose our experiment settings and implementation details in Section \ref{sec:param-setting} and Appendix\ref{appendix:details}
    
\item {\bf Code of ethics}
    \item[] Question: Does the research conducted in the paper conform, in every respect, with the NeurIPS Code of Ethics \url{https://neurips.cc/public/EthicsGuidelines}?
    \item[] Answer:\answerYes{} % Replace by \answerYes{}, \answerNo{}, or \answerNA{}.
    \item[] Justification: We have reviewed the NeurIPS Code of Ethics and confirm that our research complies with all outlined principles.

\item {\bf Broader impacts}
    \item[] Question: Does the paper discuss both potential positive societal impacts and negative societal impacts of the work performed?
    \item[] Answer: \answerYes{} % Replace by \answerYes{}, \answerNo{}, or \answerNA{}.
    \item[] Justification: This work proposes an advanced method for improving LLM's performace, and is positive for the improvement for the whole society. We discuss the impact of our work in the Introduction \ref{sec:intro}, Discussion \ref{subsec:comparison}, and Further Analysis part \ref{sec:transfer}.
    
\item {\bf Safeguards}
    \item[] Question: Does the paper describe safeguards that have been put in place for responsible release of data or models that have a high risk for misuse (e.g., pretrained language models, image generators, or scraped datasets)?
    \item[] Answer: \answerNA{} % Replace by \answerYes{}, \answerNo{}, or \answerNA{}.
    \item[] Justification: Our work does not involve the release of new models or datasets with high risk of misuse. All models and datasets used are publicly available, widely adopted in the community, and considered safe under their respective usage policies.

\item {\bf Licenses for existing assets}
    \item[] Question: Are the creators or original owners of assets (e.g., code, data, models), used in the paper, properly credited and are the license and terms of use explicitly mentioned and properly respected?
    \item[] Answer: \answerYes{} % Replace by \answerYes{}, \answerNo{}, or \answerNA{}.
    \item[] Justification: All datasets and models used in this work, such as InfoSeek, OK-VQA, EVQA, WIT, CLIP, Qwen2-VL and Qwen2.5-VL are publicly available and used in accordance with their respective licenses. Appropriate citations are provided for each asset, and no proprietary or restricted data or models were used beyond their permitted scope.

\item {\bf New assets}
    \item[] Question: Are new assets introduced in the paper well documented and is the documentation provided alongside the assets?
    \item[] Answer: \answerYes{} % Replace by \answerYes{}, \answerNo{}, or \answerNA{}.
    \item[] Justification: All of the new assets introduced in the paper, including the code and data, is well documented and will be published. 

\item {\bf Crowdsourcing and research with human subjects}
    \item[] Question: For crowdsourcing experiments and research with human subjects, does the paper include the full text of instructions given to participants and screenshots, if applicable, as well as details about compensation (if any)? 
    \item[] Answer: \answerNA{} % Replace by \answerYes{}, \answerNo{}, or \answerNA{}.
    \item[] Justification: The paper does not involve crowdsourcing nor research with human subjects.

\item {\bf Institutional review board (IRB) approvals or equivalent for research with human subjects}
    \item[] Question: Does the paper describe potential risks incurred by study participants, whether such risks were disclosed to the subjects, and whether Institutional Review Board (IRB) approvals (or an equivalent approval/review based on the requirements of your country or institution) were obtained?
    \item[] Answer: \answerNA{} % Replace by \answerYes{}, \answerNo{}, or \answerNA{}.
    \item[] Justification: The paper does not involve crowdsourcing nor research with human subjects.

\item {\bf Declaration of LLM usage}
    \item[] Question: Does the paper describe the usage of LLMs if it is an important, original, or non-standard component of the core methods in this research? Note that if the LLM is used only for writing, editing, or formatting purposes and does not impact the core methodology, scientific rigorousness, or originality of the research, declaration is not required.
    %this research? 
    \item[] Answer: \answerYes{} % Replace by \answerYes{}, \answerNo{}, or \answerNA{}.
    \item[] Justification: This work introduces a novel approach that uses Vision-Language Models (VLMs), specifically Qwen2.5-VL and Qwen2-VL, as continuous memory encoders for multimodal reasoning. The VLM itself is used to synthesize training data and generate memory embeddings. For more details, We disclose all of our experiment settings in ~\ref{sec:experimental_setup} and Appendix~\ref{appendix:details}.

\end{enumerate}
% }

\end{document}